\documentclass[lettersize,journal]{IEEEtran}
\usepackage{amsmath,amsfonts}
\usepackage{array}
\usepackage[caption=false,font=normalsize,labelfont=sf,textfont=sf]{subfig}
\usepackage{textcomp}
\usepackage{stfloats}
\usepackage{url}
\usepackage{verbatim}
\usepackage{graphicx}
\usepackage{cite}

\usepackage[ruled,vlined]{algorithm2e}
\usepackage{booktabs}
\usepackage{array}
\usepackage{multirow}
\newcommand{\sysname}{GSDMM+}

\begin{document}

\title{An Enhanced Model-based Approach for Short Text Clustering}

\author{Enhao Cheng,
        Shoujia Zhang,
        Jianhua Yin$^{\dag}$,~\IEEEmembership{Member,~IEEE},
        Xuemeng Song,~\IEEEmembership{Senior~Member,~IEEE},\\
        Tian Gan,~\IEEEmembership{Member,~IEEE},
        Liqiang Nie,
        ~\IEEEmembership{Senior~Member,~IEEE},
\IEEEcompsocitemizethanks{
\IEEEcompsocthanksitem $^{\dag}$Jianhua Yin is Corresponding Author.
\IEEEcompsocthanksitem 
Enhao Cheng, Shoujia Zhang, Jianhua Yin, Tian Gan, are with the School of Computer Science and Technology, Shandong University (Qingdao), Qingdao, Shandong 266237, China. (e-mail: {enhaocheng@mail.sdu.edu.cn; sjzhang@mail.sdu.edu.cn; jhyin@sdu.edu.cn; gantian@sdu.edu.cn}).
\IEEEcompsocthanksitem 
Xuemeng Song is with the Department of Data Science, City University of Hong Kong, Hong Kong, China (e-mail: sxmustc@gmail.com).
\IEEEcompsocthanksitem 
Liqiang Nie is with the School of Computer Science and Technology, Harbin Institute of Technology (Shenzhen), Shenzhen, Guangzhou 518055, China (e-mail: nieliqiang@gmail.com).
}
}

\markboth{IEEE Transactions on Knowledge and Data Engineering}%
{Shell \MakeLowercase{\textit{et al.}}: A Sample Article Using IEEEtran.cls for IEEE Journals}


\maketitle

\begin{abstract}
Short text clustering has become increasingly important with the popularity of social media like Twitter, Google+, and Facebook.
Existing methods can be broadly categorized into two paradigms: topic model-based approaches and deep representation learning-based approaches.
This task is inherently challenging due to the sparse, large-scale, and high-dimensional characteristics of the short text data. Furthermore, the computational intensity required by representation learning significantly increases the running time.
To address these issues, we propose a collapsed Gibbs Sampling algorithm for the Dirichlet Multinomial Mixture model (GSDMM), which effectively handles the sparsity and high dimensionality of short texts while identifying representative words for each cluster. Based on several aspects of GSDMM that warrant further refinement, we propose an improved approach, \sysname{}, designed to further optimize its performance. \sysname{} reduces initialization noise and adaptively adjusts word weights based on entropy, achieving fine-grained clustering that reveals more topic-related information. Additionally, strategic cluster merging is employed to refine clustering granularity, better aligning the predicted distribution with the true category distribution.
We conduct extensive experiments, comparing our methods with both classical and state-of-the-art approaches. The experimental results demonstrate the efficiency and effectiveness of our methods. 
The source code for our model is publicly available at \url{https://github.com/chehaoa/VEMC}.
\end{abstract}


\section{Introduction} \label{sec:introduction}

\IEEEPARstart{T}{he} proliferation of mobile internet has led to an exponential increase in user-generated data on online platforms, including video, text, and image data. Intelligent processing of such data can significantly enhance the quality of life across society and generate substantial economic benefits. Short text data are a prevalent and important form of user-generated data, consisting of concise texts such as microblogs and comments. This genre of data can reveal rich information about user’s intentions, emotions, interests, and other aspects. Short text clustering is a key technique for intelligent processing of short text data, which aims to group similar short texts together based on their contents or semantics. Short text clustering has many application scenarios, such as information retrieval~\cite{carpineto2012consensus}, event exploration~\cite{feng2015streamcube}, text summarization~\cite{shou2013sumblr}, and content recommendation~\cite{pazzani2007content}. 
Berkhin~\cite{Clustering2006survey} has discussed several important issues of clustering: 1) Setting of the number of clusters; 2) Ability to work with high-dimensional data; 3) Interpretability of results; 4) Scalability to large datasets. Short text clustering has all the above challenges. Different from normal text clustering, short text clustering also suffers from data sparsity~\cite{ZhaiTextmining2012} due to the limited words. 
Traditional clustering algorithms, such as K-means~\cite{dhillon2001concept} and hierarchical agglomerative clustering~\cite{mullner2011modern}, have been widely applied to various text clustering tasks. However, these approaches often rely on the Vector Space Model (VSM)~\cite{salton1975VSM}, where each word dimension is treated equally and independence is assumed among features. Furthermore, if we use the vector space model to represent the short texts, the sparse and high-dimensional vectors will result in waste of both memory and computation time.

Specifically, we try to cope with the above challenges of short text clustering and propose a collapsed Gibbs Sampling algorithm for the Dirichlet Multinomial Mixture (DMM) model~\cite{nigam2000text} for short text clustering (abbr. to GSDMM). GSDMM sets an upper bound on the number of clusters and automatically prunes redundant clusters during Gibbs sampling. The gradual reduction in the number of clusters resembles hierarchical clustering, enabling GSDMM to effectively uncover deep thematic structures. Through mathematical derivation, we explore how and why GSDMM works as well as the meaning of its parameters.

We find that GSDMM has the following nice properties: 1) GSDMM can infer the number of clusters automatically; 
2) Unlike the VSM-based approaches, GSDMM can cope with the sparse and high-dimensional problem of short texts; 3) Like topic models~\cite{Blei:LDA2003}, GSDMM can also obtain the representative words of each cluster. This enables GSDMM to efficiently address key clustering challenges, including the sparsity issue in short texts.

Next, although the proposed GSDMM demonstrates outstanding performance\cite{GSDMM}, an in-depth analysis reveals several aspects that warrant further refinement.

1) GSDMM randomly assigns documents to clusters during initialization, which often introduces noise and leads to instability in the early stages of sampling. 

2) The DMM model assumes a uniform Dirichlet prior for word distributions, thereby assigning equal importance to all words. This oversimplification can diminish the impact of highly discriminative words while overemphasizing globally frequent but less informative terms. 

3) Although GSDMM can prune redundant clusters within a limited number of iterations, it still risks producing a suboptimal number of clusters. Moreover, we also need to consider how to optimize GSDMM when the number of categories is known to achieve better performance. 

Based on these findings, we propose three key enhancements to the original GSDMM, thereby introducing \sysname{}, a refined approach designed to further optimize its performance.

1) A new adaptive clustering initialization introduces less noise into clustering because it assigns each document to a cluster according to the probability of the document belonging to each cluster. This not only accelerates convergence but also provides more semantically coherent starting clusters.

2) We introduce entropy-based word weighting to adaptively adjust the weight of different words based on their informational significance within each cluster. This highlights discriminative words while reducing the impact of less informative terms.

3) We dynamically adjust the clustering granularity through cluster merging to better align the predicted distribution with the true category distribution. This effectively consolidates redundant clusters and enhances clustering performance.

The contributions of our work can be summarized as follows: 
\begin{enumerate} 
    \item We propose GSDMM that is an attempt to apply the DMM model to short text clustering, and our experimental study has validated its effectiveness. We find it can cope with the sparse and high-dimensional problem of short texts, and can obtain the representative words of each cluster. Specifically, we propose GSDMM, a collapsed Gibbs sampling algorithm for the DMM model, which demonstrates high performance in short text clustering.
    \item We propose \sysname{}, an enhanced GSDMM method consisting of three modules: adaptive clustering initialization, entropy-based word weighting, and granularity adjustment with cluster merging. This method reduces initial noise and adaptively adjusts word weights based on entropy, achieving fine-grained clustering that reveals more topic-related information. Furthermore, merging clusters reduces granularity and improves clustering quality.
    \item We conduct extensive experiments, comparing our method with both classical and state-of-the-art approaches. The experimental results demonstrate the efficiency and effectiveness of our method. Furthermore, through detailed structural and hyper-parameter analysis, we explore the significance of each module and the impact of different hyper-parameters.
\end{enumerate}

The rest of this paper is organized as follows. Section~\ref{method} first introduces the DMM model and the GSDMM algorithm, then presents an improved version, \sysname{}, by enhancing the original GSDMM. Section~\ref{Discussion} discusses several important aspects of both GSDMM and GSDMM+. Section~\ref{sec:Exp} describes the experimental design used to evaluate the performance of our model and provides a detailed analysis. Section~\ref{related_works} reviews related work on text clustering. Finally, Section~\ref{sec:conclusion} presents conclusions and outlines future research directions.

\section{Method} \label{method}
In this section, we first introduce the Dirichlet Multinomial Mixture (DMM) model~\cite{nigam2000text}. Next, we provide a detailed derivation and analysis of the proposed GSDMM. Finally, we identify the limitations of GSDMM and present an enhanced model-based approach for short text clustering to address these issues.

\subsection{The DMM Model}

\begin{table}
\caption{Notations}
\centering
\begin{tabular}{cl}
  \hline
  $V$ & size of the vocabulary \\
  $D$ & number of documents in the corpus \\
  $I$ & number of iterations \\
  $\bar{L}$ & average length of documents \\
  $\vec{d}$  & documents in the corpus\\
  $\vec{z}$  & cluster assignments of each document \\
  $m_z$ & number of documents in cluster $z$ \\
  $n_z$ & number of words in cluster $z$ \\
  $n_z^{w}$   & frequency of word $w$ in cluster $z$ \\
  $N_d$   & number of words in document $d$ \\
  $N_d^{w}$   & frequency of word $w$ in document $d$ \\
  $\alpha$ & pseudo number of documents in each cluster \\
  $\beta$ & pseudo frequency of each word in each cluster \\
  \hline
\end{tabular}
\label{tab:Notations}
\end{table}

The DMM model~\cite{nigam2000text} is a probabilistic generative model for documents that is based on two key assumptions: 1) the documents are generated by a mixture model~\cite{mclachlan_basford88}, and 2) there is a one-to-one correspondence between the mixture components and the clusters. The generative process of the DMM model is illustrated as follows:
\begin{align}
    & \label{gen:dmm1}
    \theta|\alpha \sim Dir(\alpha) \\
    & \label{gen:dmm2}
    z_d|\theta \sim Mult(\theta) \qquad d=1,...,D\\
    & \label{gen:dmm3}
    \phi_k|\beta \sim Dir(\beta) \qquad k=1,...,K\\
    & \label{gen:dmm4}
    d|z_d,\{\phi_k\}_{k=1}^K \sim p(d | \phi_{z_d})
\end{align}
Here, ``$X \sim S$'' means ``$X$ is distributed according to $S$'', so the right side is a specification of distribution. When generating document $d$, the DMM model first selects a mixture component (cluster) $z_d$ for document $d$ according to the mixture weights (weights of clusters), $\theta$, in Equation~\eqref{gen:dmm2}. Then, document $d$ is generated by the selected mixture component (cluster) from distribution $p(d | \phi_{z_d})$ in Equation~\eqref{gen:dmm4}. The weight vector of the clusters, $\theta$, is generated by a Dirichlet distribution with a hyper-parameter $\alpha$, as in Equation~\eqref{gen:dmm1}. The cluster parameters $\phi_z$ are also generated by a Dirichlet distribution with a hyper-parameter $\beta$, as in Equation~\eqref{gen:dmm3}.

The probability of document $d$ generated by cluster $z_d$ is defined as follows:
\begin{equation} \label{eq:Pdz}
    p(d | \phi_{z_d}) = \prod_{w \in d} Mult(w|\phi_{z_d})
\end{equation}
Here, we make the Naive Bayes assumption: the words in a document are generated independently when the document's cluster assignment $z_d$ is known. We also assume that the probability of a word is independent of its position within the document.

\subsection{The GSDMM Algorithm} \label{GSDMM}
In this part, we give the derivation of the collapsed Gibbs sampling algorithm for the DMM model. The documents $\vec{d}=\{d_i\}_{i=1}^D$ are observed and the cluster assignments $\vec{z} = \{z_i\}_{i=1}^D$ are latent. The detail of our GSDMM algorithm is shown in Algorithm~\ref{Algo:GSDMM}, and the meaning of its variables is shown in Table~\ref{tab:Notations}. The conditional probability of document $d$ choosing cluster $z$ given the information of other documents and their cluster assignments can be factorized as follows:
\begin{align}
    & \notag
    p(z_d=z | \vec{z}_{\neg{d}}, \vec{d}, \alpha, \beta)\\
    & \label{eq:dpm1.1}
    \propto p(z_d=z | \vec{z}_{\neg{d}}, \vec{d}_{\neg{d}}, \alpha, \beta)
            p(d | z_d=z, \vec{z}_{\neg{d}}, \vec{d}_{\neg{d}}, \alpha, \beta)\\
    & \label{eq:dpm1.2}
    \propto p(z_d=z | \vec{z}_{\neg{d}}, \alpha)
            p(d | z_d=z, \vec{d}_{z, \neg{d}}, \beta)
\end{align}
Here, we use the Bayes Rule in Equation~\eqref{eq:dpm1.1} and apply the properties of D-Separation~\cite{bishop2006pattern} in Equation~\eqref{eq:dpm1.2}.

The first term in Equation~\eqref{eq:dpm1.2} indicates the probability of document $d$ choosing cluster $z$ when we know the cluster assignments of other documents. The second term in Equation~\eqref{eq:dpm1.2} can be considered as a predictive probability of document $d$ given $\vec{d}_{z, \neg{d}}$, i.e., the other documents currently assigned to cluster $z$.

The first term in Equation~\eqref{eq:dpm1.2} can be derived as follows:
\begin{align}
    & \notag
    p(z_d=z | \vec{z}_{\neg{d}}, \alpha)\\
    & \label{eq:dpm2.1}
    = \int{p(z_d=z, \theta | \vec{z}_{\neg{d}}, \alpha)
           d\theta}\\
    & \label{eq:dpm2.2}
    = \int{p(\theta | \vec{z}_{\neg{d}}, \alpha)
           p(z_d=z | \vec{z}_{\neg{d}}, \theta, \alpha)
           d\theta}\\
    & \label{eq:dpm2.3}
    = \int{p(\theta | \vec{z}_{\neg{d}}, \alpha)
           p(z_d=z | \theta)
           d\theta}\\
    &\label{eq:dpm2.4.1}
    = \int{Dir(\theta | \vec{m}_{\neg{d}} + \vec{\alpha})
           Mult(z_d=z | \theta)
           d\theta}\\
    &\label{eq:dpm2.4.2}
    = \int{ \frac{1}{\Delta(\vec{m}_{\neg{d}}+\vec{\alpha})}
        \theta_z \prod_{k=1,k\neq z}^K \theta_k^{m_{k,\neg{d}}+\alpha-1}
         d\theta}\\
    &\label{eq:dpm2.4.3}
    = \frac{\Delta(\vec{m} + \vec{\alpha})}
           {\Delta(\vec{m}_{\neg{d}} + \vec{\alpha})}\\
    &\label{eq:dpm2.4.4}
    = \frac{\prod_{k=1}^K \Gamma(m_k + \alpha)}
            {\Gamma(\sum_{k=1}^K (m_k + \alpha))}
      \frac{\Gamma(\sum_{k=1}^K (m_{k,\neg{d}} + \alpha))}
            {\prod_{k=1}^K \Gamma(m_{k,\neg{d}} + \alpha)} \\
    &\label{eq:dpm2.4.5}
    = \frac{\Gamma(m_{z,\neg{d}} + \alpha + 1)}
            {\Gamma(m_{z,\neg{d}} + \alpha)}
      \frac{\Gamma(D - 1 + K\alpha)}
            {\Gamma(D + K\alpha)} \\
    &\label{eq:dpm2.4.6}
    = \frac{m_{z,{\neg{d}}} + \alpha}
           {D - 1 + K\alpha}
\end{align}
Here, Equation~\eqref{eq:dpm2.1} exploits the sum rule of probability~\cite{bishop2006pattern}. We use the product rule of probability~\cite{bishop2006pattern} in Equation~\eqref{eq:dpm2.2} and apply the properties of D-Separation in Equation~\eqref{eq:dpm2.3}. The posterior distribution of $\theta$ is a Dirichlet distribution because we assume Dirichlet prior $Dir(\theta | \alpha)$ for the Multinomial distribution $Mult(\vec{z} | \theta)$. In Equation~\eqref{eq:dpm2.4.3}, we adopt the $\Delta$ function used in~\cite{Hei09}, which is defined as $\Delta(\vec{\alpha}) = \frac{\prod_{k=1}^K \Gamma(\alpha)}{\Gamma(\sum_{k=1}^K \alpha)}$. Using the property of $\Gamma$ function: $\Gamma(x+1)=x\Gamma(x)$, we can get Equation~\eqref{eq:dpm2.4.6} from Equation~\eqref{eq:dpm2.4.5}. In Equation~\eqref{eq:dpm2.4.6}, $m_{z,{\neg{d}}}$ is the number of documents in cluster $z$ without considering document $d$, and $D$ is the total number of documents in the dataset. Equation~\eqref{eq:dpm2.4.6} indicates that document $d$ will tend to choose larger clusters when we only consider the cluster assignments of the other documents.

\begin{algorithm}
\SetAlFnt{\small}
\DontPrintSemicolon
\KwData{Documents in the input, $\vec{d}$. }
\KwResult{Cluster labels of each document, $\vec{z}$.}
\Begin{
    Initialize $m_z, n_z$, and $n_z^w$ as zero for each cluster $z$\;
    \For{each document $d\in [1,D]$}{
        Sample a cluster for $d$: $z_d \leftarrow z \sim Multinomial(1/K)$\;
        $m_z \leftarrow m_z+1$ and $n_z \leftarrow n_z+N_d$\;
        \For{each word $w\in d$}{
            $n_z^w \leftarrow n_z^w +N_d^w$\;
        }
    }
    \For{$i \in [1,I]$}{
        \For{each document $d\in [1,D]$}{
            Record the current cluster of $d$: $z=z_d$\;
            $m_z \leftarrow m_z-1 $ and $ n_z \leftarrow n_z-N_d$\;
            \For{each word $w \in d$}{
                $n_z^w \leftarrow n_z^w- N_d^w$\;
            }
            Sample a cluster for $d$: $z_d \leftarrow z \sim p(z_d=z|\vec{z}_{\neg{d}}, \vec{d})$ (Equation~\eqref{eq:dpm3.4.7})\;
            $m_z \leftarrow m_z+1 $ and $ n_z \leftarrow n_z+N_d$\;
            \For{each word $w \in d$}{
                $n_z^w \leftarrow n_z^w+ N_d^w$\;
            }
        }
    }
}
\caption{GSDMM}
\label{Algo:GSDMM}
\end{algorithm}

Then, we derive the second term in Equation~\eqref{eq:dpm1.2} as follows:
\begin{align}
    & \notag
    p(d | z_d=z, \vec{d}_{z, \neg{d}}, \beta)\\
    & \label{eq:dpm3.1}
    = \int{p(d, \phi_z | z_d=z, \vec{d}_{z, \neg{d}}, \beta)d\phi_z}\\
    & \label{eq:dpm3.2}
    = \int{p(\phi_z | z_d=z, \vec{d}_{z, \neg{d}}, \beta)
           p(d | \phi_z, z_d=z, \vec{d}_{z, \neg{d}}, \beta)
           d\phi_z}\\
    & \label{eq:dpm3.3}
    = \int{p(\phi_z | \vec{d}_{z, \neg{d}}, \beta)
           p(d | \phi_z, z_d=z)
           d\phi_z}\\
    & \label{eq:dpm3.4.1}
    = \int{Dir(\phi_z | \vec{n}_{z,\neg{d}} + \beta)
           \prod_{w \in d} Mult(w | \phi_z)
           d\phi_z}\\
    & \label{eq:dpm3.4.2}
    = \int{\frac{1}{\Delta(\vec{n}_{z,\neg{d}} + \beta)}
            \prod_{t=1}^V \phi_{z,t}^{n_{z,\neg{d}}^t + \beta - 1}
           \prod_{w \in d} \phi_{z,w}^{n_d^w}
            d\phi_z}\\
    & \label{eq:dpm3.4.3}
    = \frac{\Delta(\vec{n}_z + \beta)}
           {\Delta(\vec{n}_{z,\neg{d}} + \beta)}\\
    & \label{eq:dpm3.4.4}
    = \frac{\prod_{t=1}^V \Gamma(n_z^t + \beta)}
            {\Gamma(\sum_{t=1}^V (n_z^t + \beta))}
      \frac{\Gamma(\sum_{t=1}^V (n_{z,\neg{d}}^t + \beta))}
            {\prod_{t=1}^V \Gamma(n_{z,\neg{d}}^t + \beta)} \\
    & \label{eq:dpm3.4.5}
    =  \frac{\prod_{w \in d} \prod_{j=1}^{N_d^w} (n_{z,\neg{d}}^w + \beta + j - 1)}
           {\prod_{i=1}^{N_d} (n_{z,\neg{d}} + V\beta + i - 1)}
\end{align}
Here, Equation~\eqref{eq:dpm3.1} exploits the sum rule of probability~\cite{bishop2006pattern}. We use the product rule of probability in Equation~\eqref{eq:dpm3.2} and apply the properties of D-Separation \cite{bishop2006pattern} to obtain Equation \eqref{eq:dpm3.3}.
The posterior distribution of $\phi_z$ is a Dirichlet distribution because we assume Dirichlet prior $Dir(\phi_z | \beta)$ for the Multinomial distribution $Mult(\vec{d}_z | \phi_z)$. Because the $\Gamma$ function has the following property: $\frac{\Gamma(x+m)}{\Gamma(x)} = \prod_{i=1}^m(x+i-1)$, we can get Equation \eqref{eq:dpm3.4.5} from Equation~\eqref{eq:dpm3.4.4}.
In Equation~\eqref{eq:dpm3.4.5}, $N_d^w$ and $N_d$ are the number of occurrences of word $w$ in document $d$ and the total number of words in document $d$, respectively, and $N_d = \sum_{w \in d} N_d^w$.
Besides, $n_{z,\neg{d}}^w$ and $n_{z,\neg{d}}$ are the number of occurrences of word $w$ in cluster $z$ and the total number of words in cluster $z$ without considering document $d$, respectively, and $n_{z,\neg{d}} = \sum_{w=1}^V n_{z,\neg{d}}^w$.
We can notice that Equation~\eqref{eq:dpm3.4.5} evaluates some similarity between document $d$ and cluster $z$, and document $d$ will tend to choose a cluster whose documents share more words with it.

Finally, we have the probability of document $d$ choosing cluster $z_d$ given the information of other documents and their cluster assignments as follows:
\begin{align}
& \notag
p(z_d=z | \vec{z}_{\neg{d}}, \vec{d}, \alpha, \beta)\\
&
\label{eq:dpm3.4.7}
\propto \frac{m_{z,\neg{d}} + \alpha}{D-1 + K\alpha}
        \frac{\prod_{w \in d} \prod_{j=1}^{N_d^w} (n_{z,\neg{d}}^w + \beta + j - 1)}
           {\prod_{i=1}^{N_d} (n_{z,\neg{d}} + V\beta + i - 1)}
\end{align}
The first part of Equation~\eqref{eq:dpm3.4.7} relates to Rule 1 of GSDMM (i.e., choose a cluster with more documents). This is also known as the ``richer get richer" property, leading larger clusters to get larger~\cite{teh2010dirichlet}. The second part of Equation~\eqref{eq:dpm3.4.7} relates to Rule 2 of GSDMM (i.e., choose a cluster that is more similar to the current document), which defines the similarity between the document and the cluster. Notably, GSDMM represents each document by its words and their respective frequencies. It models each cluster as a single large document composed of all documents within the cluster while recording the cluster size.

\subsection{The \sysname{} Algorithm} \label{VEMC}
Although the proposed GSDMM demonstrates outstanding performance, an in-depth analysis reveals several aspects that warrant further refinement: 1) Due to the lack of initial document information, random initialization leads to poor performance in early iterations and requires more iterations to correct misallocated documents. 2) When assigning documents to clusters, the uniform word weighting scheme ignores the varying contributions of different words to clusters, potentially introducing noise. 3) GSDMM assumes that at most $K_{max}$ clusters are in the corpus, which is a weaker assumption than specifying the true number of categories. Although GSDMM can prune redundant clusters within a limited number of iterations, it still risks producing a suboptimal number of clusters.

To address these issues, we propose an improved short text clustering method \sysname{} which is shown in Algorithm~\ref{Algo:VEMC}. First, we replace random initialization with adaptive clustering initialization to accelerate model convergence. Next, we incorporate entropy to measure word importance and dynamically adjust word weights, enabling the model to focus more on key terms. Finally, we dynamically refine the number of clusters to better align the predicted distribution with the true category distribution. In this section, we provide a detailed explanation of each improvement.

\begin{algorithm}[t]
\setlength{\belowcaptionskip}{-0.10cm}
\SetAlFnt{\scriptsize}
\DontPrintSemicolon
\KwData{Document vector $\vec{d}$, Maximum number of clusters $K_{max}$, Real number of clusters $K_{real}$.}
\KwResult{cluster assignments of each document $\vec{z}$.}
\Begin{
    //Adaptive Clustering Initialization\;
    Obtain the initial cluster assignments for documents by Algorithm \ref{Algo:ACI}.\;
    //Collapsed Gibbs Sampling with Entropy\;
    \For{$i \in [1,I]$}{
        \For{each document $d\in [1,D]$}{
            Record the current cluster of $d$: $z=z_d$\;
            $m_z \leftarrow m_z-1 $ and $ n_z \leftarrow n_z-N_d$\;
            \For{each word $w \in d$}{
                $n_z^w \leftarrow n_z^w- N_d^w$\;
            }
            \If{$n_z == 0$}{
                //Remove the empty cluster\;
                $K \leftarrow K-1$\;
                Re-arrange cluster indices so that $1,...,K$ are active (i.e., non-empty);\;
            }
            \If{d \% 15 == 0}{
                Update the word entropy values.\;
            }
            Compute the probability of document $d$ and choose a new $z$ from each of the $K$ existing clusters with Equation \eqref{DMM-Entropy}.\;
            $z_d \leftarrow z$\;
            $m_z \leftarrow m_z+1 $ and $ n_z \leftarrow n_z+N_d$\;
            \For{each word $w \in d$}{
                $n_z^w \leftarrow n_z^w+ N_d^w$\;
            }
        }
    }
    //Granularity Adjustment with Cluster Merging\;
    \If{$K > K_{\text{real}}$}{
        Compute the TF-ICF word distributions of clusters by Equation \eqref{idf} and Equation \eqref{tfidf}.\;
        Build a priority queue $Q$ of cluster pairs.\;
        \While{$K > K_{\text{real}}$}{
            //Highest-similarity pair\;
            $(c_p, c_q) \leftarrow \text{PopMax}(Q)$\;
            $m_{c_m} \leftarrow m_{c_p} + m_{c_q}$\;
            $n_{c_m} \leftarrow n_{c_p} + n_{c_q}$\;
            \For{each word $w$}{
                $n_{c_m}^w \leftarrow n_{c_p}^w + n_{c_q}^w$\;
            }
            $K \leftarrow K - 1$\;
            \For{each active cluster $c_a$}{
                Compute the cosine similarity of $(c_m, c_a)$ and insert it into $Q$.\;
            }
        }
    }
}
\caption{\sysname{}}
\label{Algo:VEMC}
\end{algorithm}

\subsubsection{Adaptive Clustering Initialization}
Traditional random initialization assigns a random label to each document, often ignoring distance information between documents. This results in poor initial performance and requires more iterations to achieve stability. Therefore, we introduce a new initialization method, presented in Algorithm~\ref{Algo:ACI}, which incorporates the similarity of the document cluster in the initialization stage to improve the effectiveness.


The document set is initially composed of $K$ randomly selected documents, each serving as an initial cluster. Unlike K-means and the standard Voronoi diagram~\cite{aurenhammer2000voronoi}, which assign each data point to the nearest fixed cluster center, our method iteratively refines clusters during initialization. Each document is assigned to its most similar cluster, and the clusters are continuously updated based on the assigned documents. This process provides a more interpretable and stable starting point for subsequent clustering optimization.


\subsubsection{Entropy-based Word Weighting}
The standard Dirichlet multinomial mixture model used in GSDMM assumes an uniform Dirichlet prior parameter ($\beta$) for all words. This assumption implies that all words contribute equally to cluster formation, oversimplifying the nature of textual data. In reality, words exhibit varying levels of discriminative power across categories. 

To address this issue, we introduce entropy-based word weighting to adaptively adjust the influence of individual words based on their informational significance within each cluster.
For each word $w \in V$, its entropy $H(w)$ across the $K$ clusters is computed as follows:
\begin{align}
& \label{pk}
p_k(w) = \frac{n_k^w+\epsilon}{\sum_{k=1}^K n_k^w+K\epsilon} \\
& \label{hw}
H(w) = - \sum_{k=1}^K p_k(w) \log p_k(w)
\end{align}
where $p_k(w)$ represents the proportion of word $w$ in cluster $k$ relative to its total occurrences across all clusters, estimated based on the current cluster assignments. A small value $\epsilon$ is used to prevent division by zero errors. The entropy $H(w)$ measures the uncertainty in the word's distribution across clusters. Words with lower entropy are more informative, as they are predominantly associated with specific clusters. Conversely, words with higher entropy are more evenly distributed across clusters, making them less useful for clustering.

To incorporate word weighting based on entropy, we utilize the entropy of each word to replace the hyper-parameter $\beta$, which adjusts the influence of each word. This approach allows us to compute the probability of document $d$ selecting cluster $z$ as follows:
\begin{align}
& \notag
p(z_d=z | \vec{z}_{\neg{d}}, \vec{d}, \alpha)\\
&
\label{DMM-Entropy}
\propto \frac{m_{z,\neg{d}} + \alpha}{D-1 + K\alpha}
        \frac{\prod_{w \in d} \prod_{j=1}^{N_d^w} (n_{z,\neg{d}}^w + H(w) + j - 1)}
           {\prod_{i=1}^{N_d} (n_{z,\neg{d}} + \sum_{w \in V} H(w) + i - 1)}
\end{align}

This adjustment ensures that Gibbs sampling accounts for the relative importance of each word based on its entropy, allowing the model to prioritize words with higher discriminative power during cluster assignment. Specifically, in Equation~\eqref{DMM-Entropy}, words with higher entropy reduce the influence of the word's frequency in cluster $z$, thereby diminishing their impact on clustering. In contrast, words with lower entropy increase the effect of the frequency $n_{z,\neg{d}}^w$, enhancing their distinguishing power.

This approach enhances the influence of discriminative words while mitigating the noise, allowing for a greater focus on meaningful terms. By prioritizing words that contribute higher mutual information, it aligns with information-theoretic principles and strengthens the model's ability to distinguish clusters. Furthermore, by dynamically adjusting word weights based on entropy, \sysname{} significantly improves upon GSDMM and enhances clustering accuracy, particularly for complex textual datasets.

\begin{algorithm}[t]
\setlength{\belowcaptionskip}{-0.10cm}
\SetAlFnt{\scriptsize}
\DontPrintSemicolon
\KwData{Document vector $\vec{d}$, maximum number of clusters $K_{max}$.}
\KwResult{Initial cluster assignments of each document $\vec{z}$.}
\Begin{
    \For{each cluster $z \in [1, K_{max}]$}{
        Sample a document $d$ with a uniform distribution.\;
        $z_d \leftarrow z$\;
        $m_z \leftarrow m_z+1 $ and $ n_z \leftarrow n_z+N_d$\;
        \For{each word $w \in d$}{
            $n_z^w \leftarrow n_z^w+ N_d^w$\;
        }}
    Construct an empty document set $S$. \;
    Add the sampled documents into $S$. \;
    \For{each document $d \in [1, D]$}{
        \If{$d$ is not in $S$} {
            Sample a new cluster $z$ for $d$: \;  
            $z_d \leftarrow z \sim p(z_d = z| \vec{z}, \vec{d}, \alpha, \beta)$ \;($\vec{d}$ represents the documents and $\vec{z}$ denotes the corresponding cluster assignments in $S$.) \;
            Add the document $d$ into $S$. \;
            $m_z \leftarrow m_z + 1$ and $n_z \leftarrow n_z + N_d$\;
            \For{each word $w \in d$}{
                $n_z^w \leftarrow n_z^w + N_d^w$\;
            }
        }
    }
}
\caption{Adaptive Clustering Initialization}
\label{Algo:ACI}
\end{algorithm}

\subsubsection{Granularity Adjustment with Cluster Merging}
In the previous modules, we generate a fine-grained cluster distribution, also known as high granularity. This process produces more small clusters, capturing subtle local patterns during the iterations. We dynamically adjust clustering granularity through cluster merging to better align the predicted distribution with the true category distribution, effectively consolidating redundant clusters and enhancing clustering performance. In this section, we design a modified Term Frequency-Inverse Document Frequency (TF-IDF) measure, called Term Frequency-Inverse Cluster Frequency (TF-ICF), which uses the weight $W_{t,c}$ to represent a term’s importance to a topic $c$. 
\begin{align}
&\label{idf}
    icf_{t} = 1 + \log \left(\frac{1 + K}{1 + cf(t)} \right) \\
&\label{tfidf}
    W_{t,c} = tf_{t,c} \cdot icf_{t}
\end{align}
where $K$ is the total number of clusters, term $t$ represents a word, and the cluster frequency $cf(t)$ is the number of clusters that contain the term $t$. The inverse cluster frequency ($icf_{t}$) measures how much information a term $t$ provides to a cluster. Then, the term frequency ($tf_{t,c}$) models the frequency of term $t$ in a cluster $c$, and the cluster $c$ is a single large document composed of all documents within the cluster while recording the cluster size. TF-ICF models the importance of words in clusters instead of individual documents, which allows us to generate topic-word distributions for each cluster of documents. 

Central to our method is the quantification of cosine similarity between clusters based on their topic-word distributions, making it well-suited for comparing semantic similarity. Specifically, to optimize the merging process and reduce computational overhead, we employ a priority queue that stores cluster pairs sorted by similarity scores. Finally, by iteratively merging the TF-ICF representations of the most semantically coherent clusters, we can reduce the number of topics to a user-specified value. This approach brings the resulting number of clusters closer to the true number of categories, yielding optimal results.


\section{Discussion}  \label{Discussion}
\subsection{Meaning of Alpha and Beta} \label{sec:disAlpahAndBeta}
In this part, we try to explore the meaning of $\alpha$ and $\beta$ in GSDMM and \sysname{}. From Equation~\eqref{eq:dpm3.4.7} and Equation~\eqref{DMM-Entropy}, we can see that $\alpha$ relates to the prior probability of a document choosing a cluster. If we set $\alpha=0$, a cluster will never be chosen by the document once it gets empty, as the first part of the two equations becomes zero. When $\alpha$ gets larger, the probability of a document choosing an empty cluster also gets larger.
In GSDMM, we can see $\beta$ is in the second part of Equation~\eqref{eq:dpm3.4.7} which relates to Rule 2. If we set $\beta=0$, a document will never choose a cluster. We can see this is not reasonable, because other words of the document may appear many times in that cluster and it may be similar to the documents of that cluster.
In \sysname{}, we observe that $\beta$ is used in the adaptive clustering initialization, which controls the number of clusters obtained prior to granularity adjustment. When $\beta$ takes small values, a fine-grained clustering is achieved, producing more clusters that capture subtle local patterns during the iterations. However, as $\beta$ increases, the number of clusters decreases, resulting in coarser granularity. Similar to GSDMM, when $\beta=0$, the model cannot be trained. Detailed validation is provided in the Section~\ref{Hyper-parameter_Analysis}.

DMM assumes symmetric priors for the Dirichlet distributions, in other words, it gives identical $\alpha$ for all clusters and consistent $\beta$'s for all words. The identical $\alpha$ for all clusters implies that different clusters are equally important at the start, which aligns with our intuition. 
In GSDMM, assigning the same $\beta$ to all words implies that all words are equally important. This is not reasonable and may mislead the clustering algorithm. We should place less emphasis on overly common words that appear in too many documents and focus more on key terms. Therefore, in \sysname{}, we achieve this goal by employing the entropy-based word weighting to adaptively adjust the weight of different words based on their informational significance within each cluster.

\subsection{Relationship with Naive Bayes Classifier} \label{sec:NBC}

The conditional distribution $p(z_d=z|\vec{z}_{\neg{d}}, \vec{d}, \alpha, \beta)$ in Equation~\eqref{eq:dpm3.4.7} is equivalent to the Bayesian Naive Bayes Classifier (BNBC)~\cite{rennie01BayesianNBC}. Intuitively, we can assign document $d$ to a cluster $z$ with the largest conditional probability $p(z_d=z|\vec{z}_{\neg{d}}, \vec{d}, \alpha, \beta)$. However, our methods choose to sample cluster $z$ from the conditional distribution in Equation~\eqref{eq:dpm3.4.7} and Equation~\eqref{DMM-Entropy}. GSDMM and \sysname{} can avoid falling into a local minimum which is a common problem of EM-based algorithms in this way.

Furthermore, Rennie \cite{rennie01BayesianNBC} points out that BNBC performs worse in classification, because it over-emphasizes words that appear more than once in a test document. For example, if word $w$ appears twice in a document $d$, the contribution of $w$ for Equation~\eqref{eq:dpm3.4.7} is $(n_{z,\neg{d}} + \beta)(n_{z,\neg{d}} + \beta + 1)$. However, this is a good property in the text clustering problem, because words in a document tend to appear in bursts: if a word appears once, it is more likely to appear again~\cite{DCMmadsen2005modeling,rennie2003Burstness}. The conditional distribution $p(z_d=z|\vec{z}_{\neg{d}}, \vec{d}, \alpha, \beta)$ in Equation~\eqref{eq:dpm3.4.7} can give words that appear multi-times in a document more emphasis, and allows GSDMM to capture the burstiness of words~\cite{DCMelkan2006clustering}. \sysname{} retains all the advantages of GSDMM while addressing its existing issues.

\subsection{Representation of Clusters}
From Algorithm \ref{Algo:GSDMM}, we can see that GSDMM can assign each document to a cluster. With the fact that the Dirichlet distribution is conjugate to the multinomial distribution, we have:
\begin{equation}\label{eq:posterior}
    p(\phi_{z} | \vec{d}, \vec{z}, \vec{\alpha}, \vec{\beta})
    =Dir(\phi_{z} | \vec{n}_z + \vec{\beta})
\end{equation}
where $\vec{n}_z = \{n_z^{w}\}_{w=1}^V$, and $n_z^{w}$ is the number of occurrences of word $w$ in the $z$th cluster.

From Equation \ref{eq:posterior}, we can obtain the posterior mean of $\phi$ as follows:
\begin{equation}\label{eq:posterior2}
    \hat{\phi}_{z,w} = \frac{n_z^{w} + \beta}{\sum_{w=1}^V n_z^{w} +V\beta}
\end{equation}
where $\hat{\phi}_{z,w}$ corresponds to the probability of word $w$ being generated by cluster $z$, and can be regarded as the importance of word $w$ to cluster $z$.
As a result, GSDMM and \sysname{} can obtain the representative words of each cluster like topic models~\cite{Blei:LDA2003}. It is interesting to notice that Equation~\eqref{eq:posterior2} actually defines the fraction of occurrences of word $w$ in cluster $z$, and is highly related to the second part of Equation~\eqref{eq:dpm3.4.7} and Equation~\eqref{DMM-Entropy}. If word $w$ has a relatively high value of $\hat{\phi}_{z,w}$, it can be regarded as the representative word of cluster $z$. In Section~\ref{Visualization_Analysis}, we present the top ten representative words for each cluster that \sysname{} finds on a dataset, and find that these words can perfectly represent those clusters.

\section{Experiment and Analysis} \label{sec:Exp}
In this section, we first introduce the experimental setup and compare \sysname{} with other state-of-the-art models to demonstrate its superiority. Next, we conduct a structural analysis to examine the functional roles of each module and validate its effectiveness. Besides, we conduct a running time analysis to evaluate efficiency and a hyper-parameter experiment to assess sensitivity.
\subsection{Experiment Setup}
\subsubsection{Dataset}
For the clustering task, we assess the performance of the proposed model on benchmark datasets for short text clustering. Table~\ref{tab:text_stats} summarizes the statistical details of these datasets.
The three News datasets are derived from Google News\footnote{\url{https://news.google.com}}, with each instance including a title and a snippet. In the Google News, the news articles are grouped into clusters (stories) automatically. Specifically, the News-T dataset contains only titles, the News-S dataset includes only snippets, and the News-TS dataset combines titles and snippets. We manually examine this dataset, and find that it is with really good quality (Almost all articles in the same cluster are about the same event, and articles in different clusters are about different events). The Tweet dataset originates from the Text Retrieval Conference\footnote{\url{https://trec.nist.gov/data/microblog.html}}. 

All datasets undergo the following preprocessing steps: (1) Convert letters into lowercase; (2) Remove non-latin characters and stop words; (3) Perform stemming for words with the WordNet Lemmatizer of NLTK\footnote{\url{http://www.nltk.org}}; (4) Remove words whose length are smaller than 2 or larger than 15; (5) Remove words with document frequency less than 2.

\begin{table}[h]
\centering
\renewcommand{\arraystretch}{1.2} 
\caption{Statistics of the short text datasets}
\label{tab:text_stats}
\begin{tabular}{lcccc}
\hline
\textbf{Dataset} & \boldmath$K$ & \boldmath$D$ & \textbf{Len} & \boldmath$V$ \\ \hline
Tweet   & 89  & 2472  & 8.55/20   & 5098   \\
News-T  & 152 & 11109 & 6.23/14   & 8110   \\
News-S  & 152 & 11109 & 21.81/33  & 18325  \\
News-TS & 152 & 11109 & 27.95/43  & 19508  \\ \hline
\end{tabular}
\end{table}

\subsubsection{Implementation Detail}
We implement the GSDMM and \sysname{} using PyTorch, and all experiments are conducted on a single NVIDIA A100 GPU. In GSDMM, $\alpha$ and $\beta$ are set to 0.1. The parameters for \sysname{} are set as $\alpha$ = 0.1 and $\beta$ = 0.01. The predefined number of clusters, $K_{max}$, is set to 500 for all datasets, while $K_{real}$ corresponds to each dataset's known number of categories. To ensure model efficiency, we set the entropy calculation frequency to 15, meaning that entropy is computed once every $D/15$ iteration, where $D$ represents the number of documents in the corpus. The number of iterations is set to 20, and we conduct 10 independent experiments for each dataset.

\subsubsection{Compared Baseline} We compare GSDMM and \sysname{} with the following state-of-the-art methods.
\begin{itemize}
    \item K-means is a conventional text clustering method that takes feature vectors of texts as input and outputs $K$ clusters. We employ two types of feature vectors: TF-IDF and text embeddings derived from SimCSE \cite{gao2021simcse}, referred to as K-means (TF-IDF) and K-means (Embedding), respectively.
    \item HieClu is another traditional clustering method. Similar to K-means, we also use two feature vectors, denoted as HieClu(TF-IDF) and HieClu(Embedding).
    \item LDA~\cite{blei2003latent} is a widely used topic modeling technique that assumes each text is generated from multiple topics.
    \item BTM~\cite{yan2013biterm} learns topics by directly modeling the generation of biterms in the dataset.
    \item WNTM~\cite{zuo2016word} constructs a word network to infer the topic distribution for each word.
    \item PTM~\cite{zuo2016topic} introduces the concept of pseudo-documents to aggregate short texts implicitly.
    \item LCTM~\cite{hu2016latent} models each topic as a distribution over latent concepts, with each latent concept being a localized Gaussian distribution in the word embedding space.
    \item SCCL~\cite{SCCL} uses the Sentence-BERT model and fine-tunes it using the contrastive loss and the clustering loss.
    \item DACL~\cite{li2022clustering} improves SCCL from the perspective of pseudo-label.
    \item RSTC~\cite{zheng2023robust} proposes self-adaptive optimal transport and introduces both class-wise and instance-wise contrastive learning.
    \item MVC~\cite{lu2024multi} models the bag-of-words view using the DMM model and the document embedding view using the GMM.
\end{itemize}

\subsection{Performance Comparison}

\begin{table*}[t]
\centering
\renewcommand{\arraystretch}{1.2} 
\setlength{\tabcolsep}{9pt} 
\caption{Performance comparison over all methods. The bold and underlined values indicate the best and runner-up results, respectively.}
\begin{tabular}{lcccccccc}
\hline
          & \multicolumn{2}{c}{\textbf{Tweet}} & \multicolumn{2}{c}{\textbf{News-T}} & \multicolumn{2}{c}{\textbf{News-S}} & \multicolumn{2}{c}{\textbf{News-TS}} \\ \cline{2-9} 
                         & \textbf{ACC} & \textbf{NMI}       & \textbf{ACC} & \textbf{NMI}       & \textbf{ACC} & \textbf{NMI}       & \textbf{ACC} & \textbf{NMI}        \\ \hline
K-means(TF-IDF)          & 0.563        & 0.795              & 0.562        & 0.777              & 0.621        & 0.831              & 0.684        & 0.879               \\
K-means(Embedding)       & 0.581        & 0.839              & 0.589        & 0.806              & 0.613        & 0.831              & 0.665        & 0.876               \\
HieClu(TF-IDF)          & 0.589        & 0.784              & 0.593        & 0.773               & 0.702        & 0.845              & 0.788        & 0.901               \\
K-means(Embedding)       & 0.628        & 0.857              & 0.641        & 0.823               & 0.681        & 0.845              & 0.756        & 0.895               \\
LDA                      & 0.609        & 0.805              & 0.680        & 0.824              & 0.769        & 0.871              & 0.801        & 0.908               \\
BTM                      & 0.575        & 0.807              & 0.715        & 0.875              & 0.738        & 0.891              & 0.762        & 0.920               \\
WNTM                     & 0.678        & 0.851              & 0.781        & 0.878              & 0.743        & 0.863              & 0.814        & 0.918               \\
PTM                      & 0.685        & 0.840              & 0.738        & 0.866              & 0.785        & 0.878              & 0.814        & 0.911               \\
LCTM                     & 0.614        & 0.797              & 0.636        & 0.784              & 0.726        & 0.863              & 0.791        & 0.894               \\
SCCL                     & 0.782        & 0.892              & 0.758        & 0.883              & 0.831        & 0.904              & \textbf{0.898}        & \underline{0.949}               \\ 
DACL                     & \underline{0.820}         & 0.906              &0.805        & 0.898               & 0.840        & \underline{0.917}              & \underline{0.896}         & 0.944               \\ 
RSTC                     & 0.752        & 0.873              & 0.722        & 0.873               & 0.793        & 0.894              & 0.832        & 0.931               \\ 
MVC                      &0.807         & \underline{0.910}              & \textbf{0.847}        & \textbf{0.918}     
        & \underline{0.845}        & 0.916              & 0.878        & 0.946               \\ \hline
GSDMM                    & 0.778        & 0.895             & 0.778        & 0.892              & 0.797       & 0.901              & 0.829        & 0.934               \\
\sysname{}   & \textbf{0.870} & \textbf{0.914}     & \underline{0.830} & \underline{0.906}    & \textbf{0.855} & \textbf{0.929}     & 0.889 & \textbf{0.957}      \\ \hline
\end{tabular}
\label{tab:performance_comparison}
\end{table*}

Table~\ref{tab:performance_comparison} compares the proposed GSDMM and \sysname{} against a range of baseline methods across four datasets: Tweet, News-T, News-S, and News-TS. Accuracy (ACC) and Normalized Mutual Information (NMI) are the evaluation metrics employed, which collectively assess the clustering performance from both classification and information-theoretic perspectives.

From the table, we can infer the following observations: 
1) GSDMM outperforms most methods while maintaining simplicity, elegance, and high efficiency. With enhancements, \sysname{} achieves significant performance improvements across all aspects.
2) \sysname{} consistently outperforms all baseline methods regarding ACC and NMI on the Tweet and News-S datasets. Notably, \sysname{} achieves the best clustering performance without leveraging text embeddings, demonstrating its superiority. In particular, we improve accuracy by 5\% on the Tweet dataset.
3) Compared to MVC, which performs text clustering using both the bag-of-words and text embedding views, our methods relies solely on the bag-of-words representation. This design ensures extremely high efficiency while still achieving competitive performance. For the News-T dataset, \sysname{} achieves the second-best performance, with an ACC of 0.83 and an NMI of 0.906, while maintaining high efficiency. 
4) \sysname{} demonstrates superior performance compared to SCCL, which uses the sentence-bert model and fine-tunes it with contrastive loss and clustering loss. The method exploits word co-occurrences in texts and is highly competitive compared to embedding-based models, thereby providing a significant advancement over existing methods.
 
\subsection{Structure Analysis}

\begin{figure*}[t]  
    \centering
    \begin{minipage}[t]{0.23\textwidth}
        \centering
        \includegraphics[width=\linewidth]{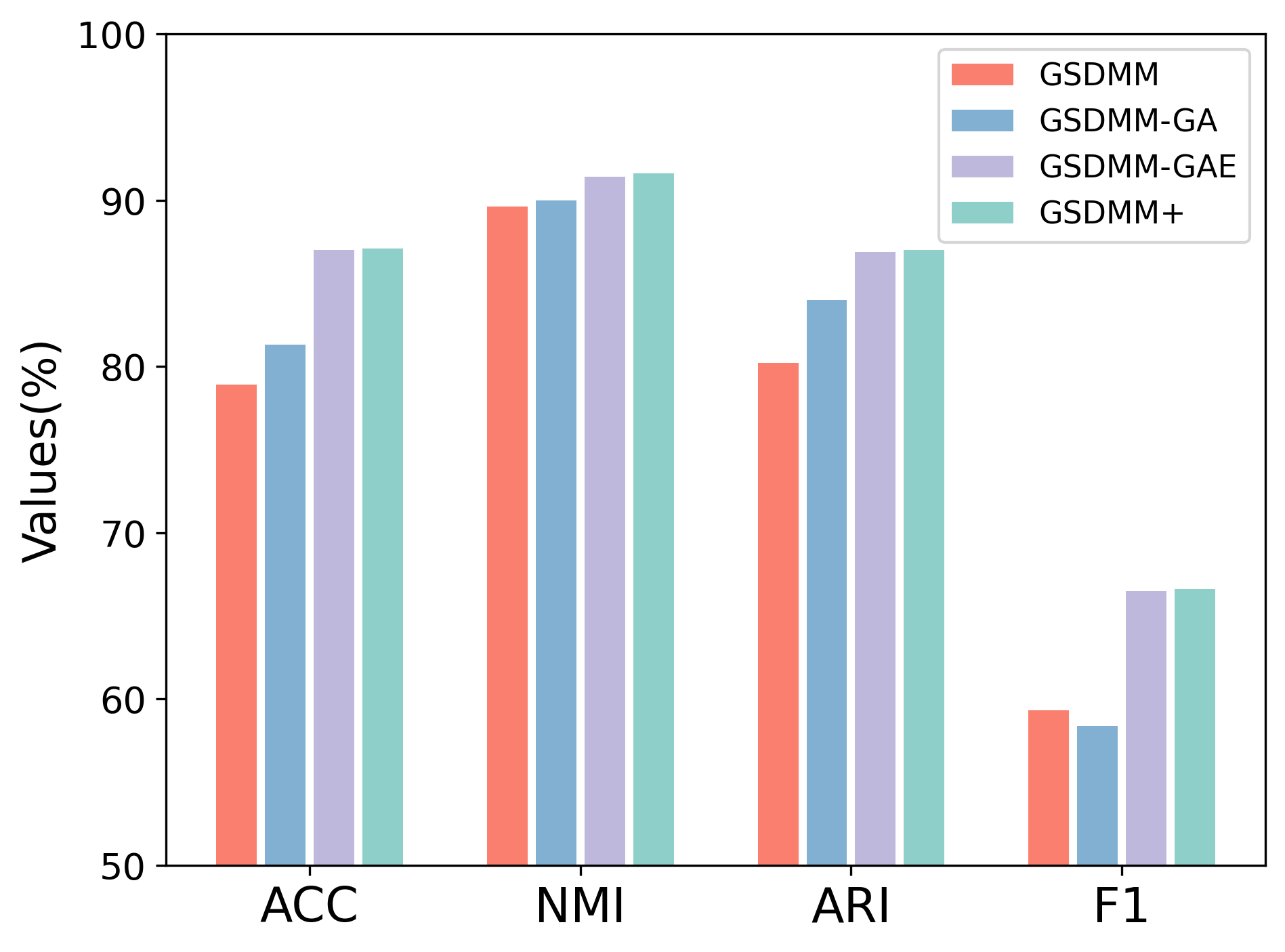}
        \centerline{(a) Tweet}
    \end{minipage}
    \hfill
    \begin{minipage}[t]{0.23\textwidth}
        \centering
        \includegraphics[width=\linewidth]{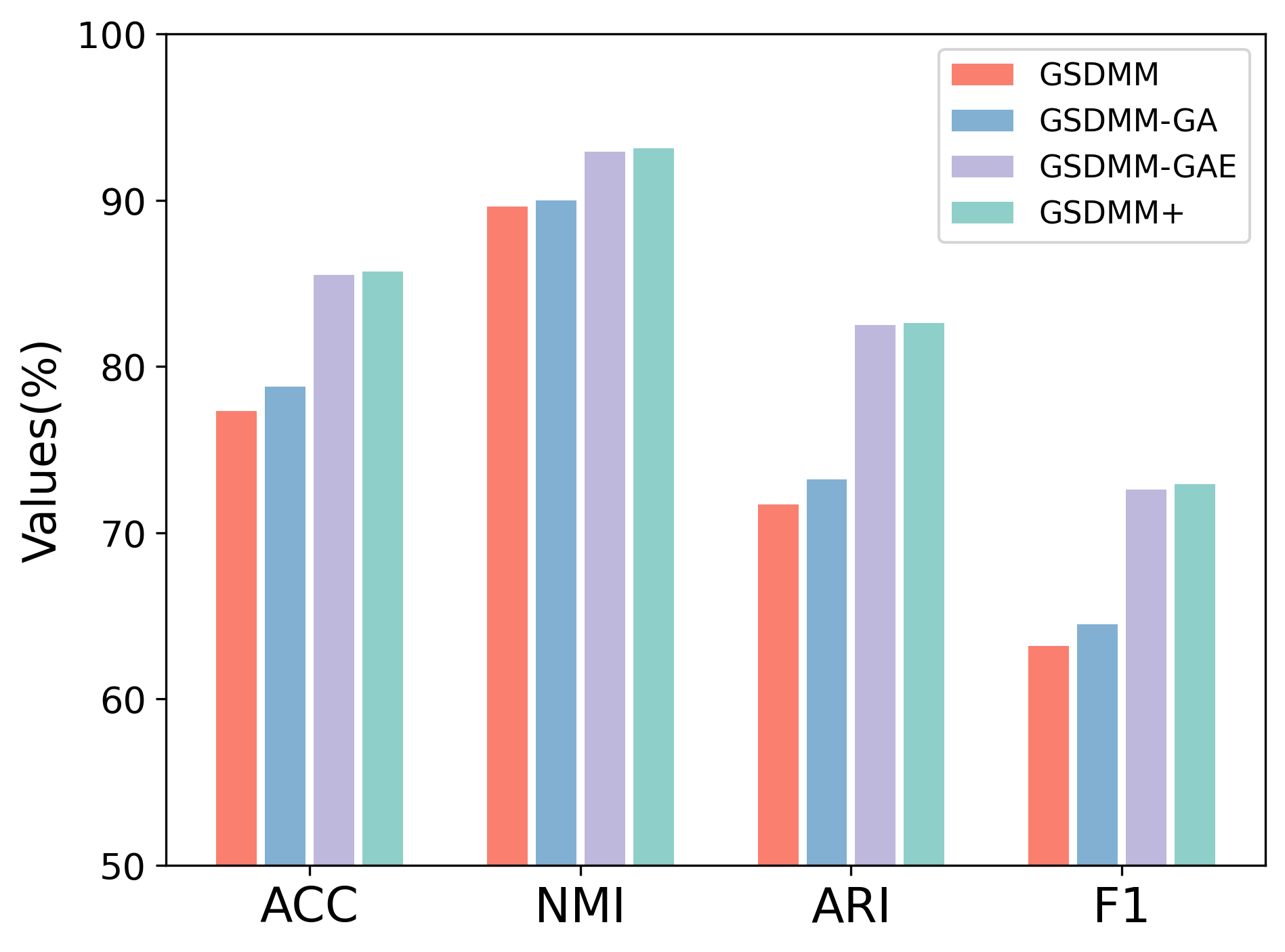}
        \centerline{(b) News-S}
    \end{minipage}
    \hfill
    \begin{minipage}[t]{0.23\textwidth}
        \centering
        \includegraphics[width=\linewidth]{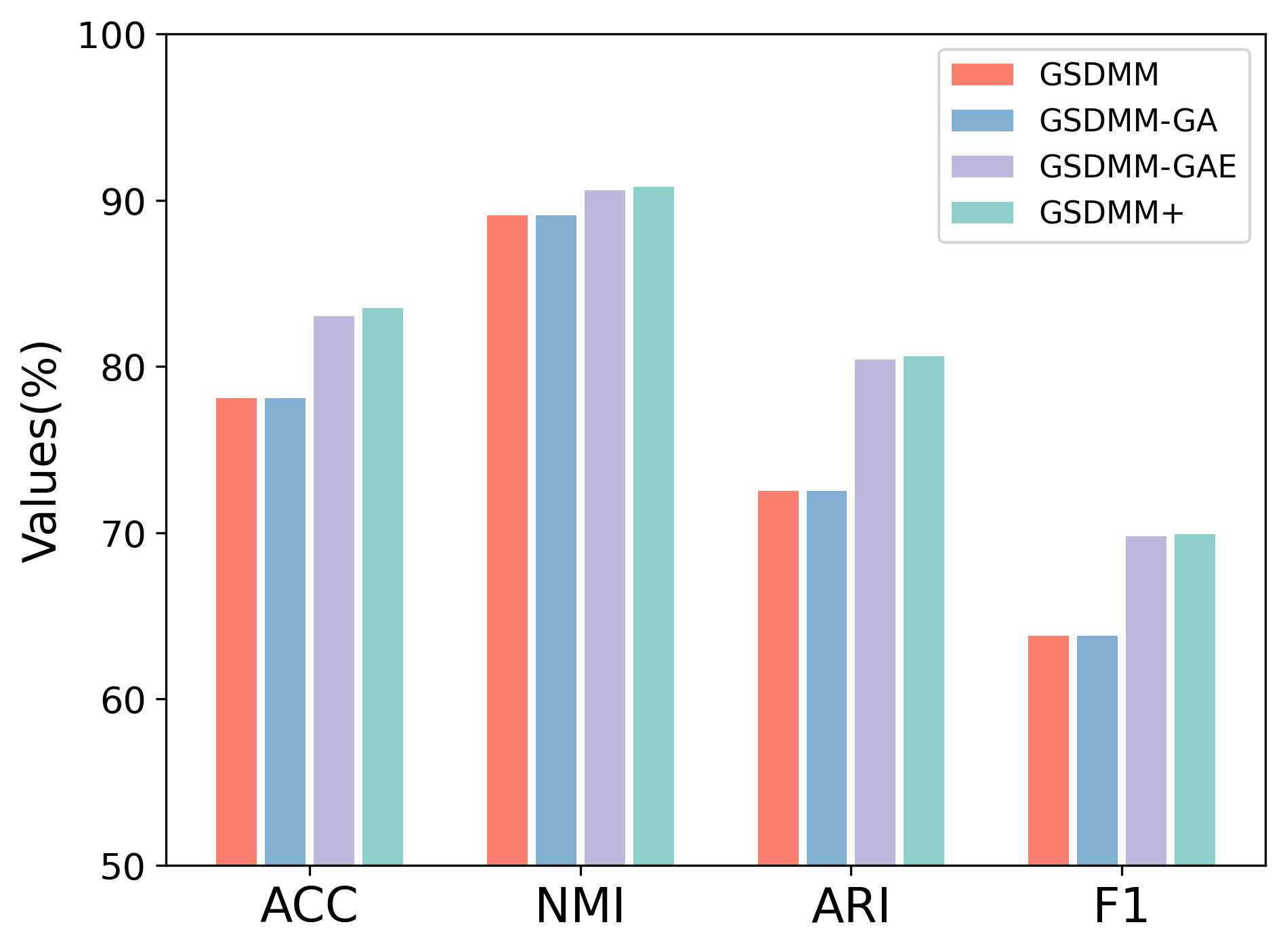}
        \centerline{(c) News-T}
    \end{minipage}
    \hfill
    \begin{minipage}[t]{0.23\textwidth}
        \centering
        \includegraphics[width=\linewidth]{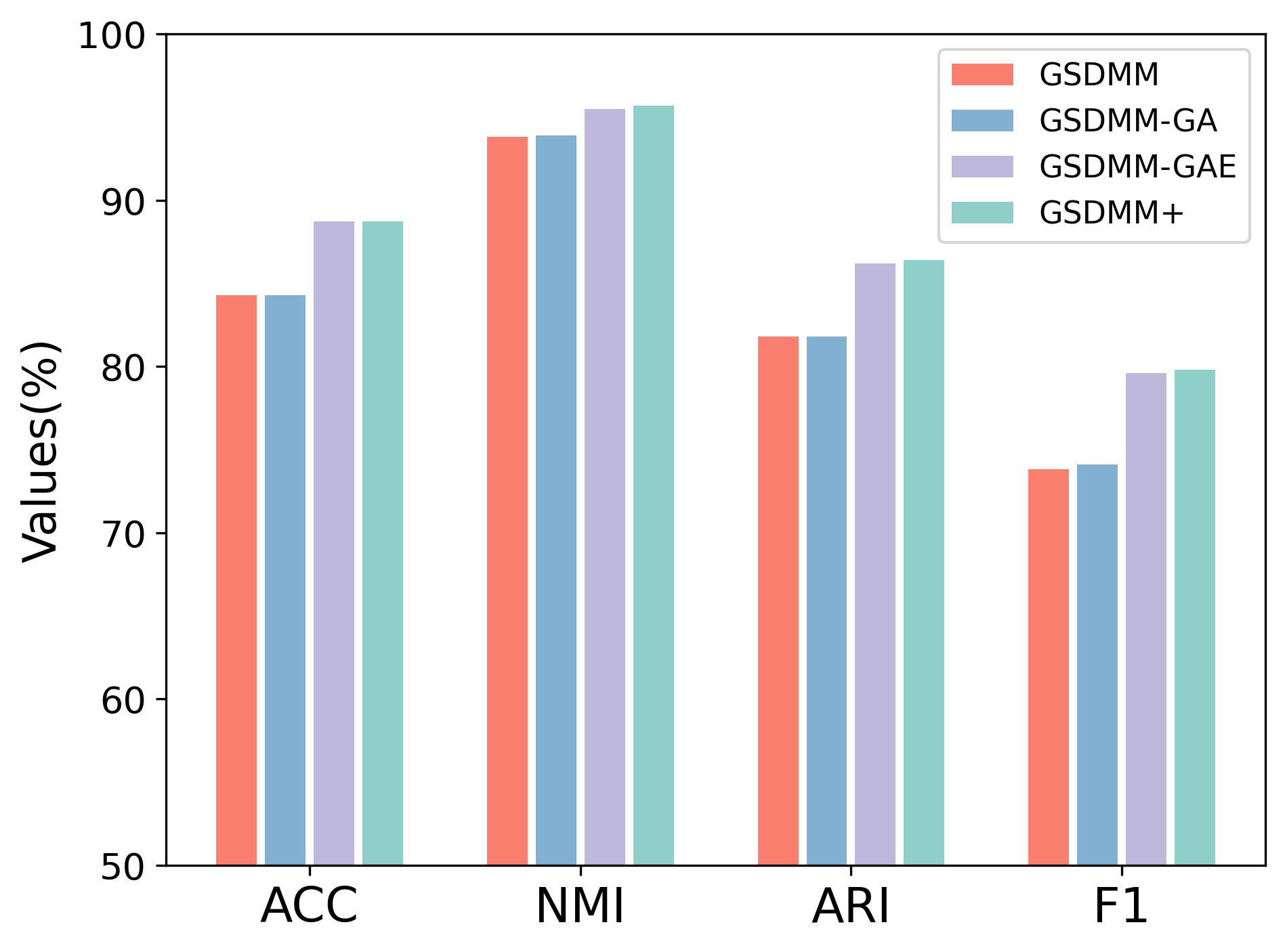}
        \centerline{(d) News-TS}
    \end{minipage}
\caption{Performance comparison of different module settings for clustering.}
\label{module_Comparison}
\end{figure*}

This section delves into how the proposed designs adjust cluster granularity and improve clustering performance in \sysname{}. Figure~\ref{module_Comparison} presents a comparison of clustering performance under four module settings:
\begin{itemize}
    \item GSDMM: The collapsed Gibbs sampling algorithm for the Dirichlet Multinomial Mixture model.
    \item GSDMM-GA: The method incorporates the granularity adjustment into GSDMM algorithm.
    \item GSDMM-GAE: The method employs entropy-based word weighting and adjusts granularity through cluster merging.
    \item \sysname{}: Our method consists of three modules: adaptive clustering initialization, entropy-based word weighting, and granularity adjustment with cluster merging.
\end{itemize}
Figure~\ref{module_Comparison} presents the ablation studies of the proposed \sysname{} on four datasets, demonstrating consistent and substantial improvements across all metrics.

From the figures, we observe that: 1) In the Tweet and News-S datasets, cluster merging enhances clustering performance by refining the number of clusters. This demonstrates the effectiveness of granularity adjustment in clustering. 2) Across all datasets, our method, which employs entropy-based word weighting and granularity adjustment with cluster merging, significantly improves clustering performance. Entropy-based word weighting refines cluster granularity, enabling the extraction of more fine-grained information. Granularity adjustment determines a reasonable number of clusters, ultimately achieving optimal performance and surpassing methods that rely on text embeddings. 3) Using only entropy-based word weighting to refine cluster granularity creates a discrepancy between the number of clusters and the true number of categories. Without fine-grained adjustment, clustering performance deteriorates significantly. Therefore, this scenario is not included in the presented figures.

\subsubsection{Adaptive Clustering Initialization}
To effectively demonstrate the necessity and effectiveness of the adaptive clustering initialization module, we present experimental results conducted on two distinct datasets: Tweet and News-TS. 

\begin{figure}[t]
\centering
\begin{minipage}{0.49\linewidth}
\centerline{\includegraphics[width=0.98\textwidth]{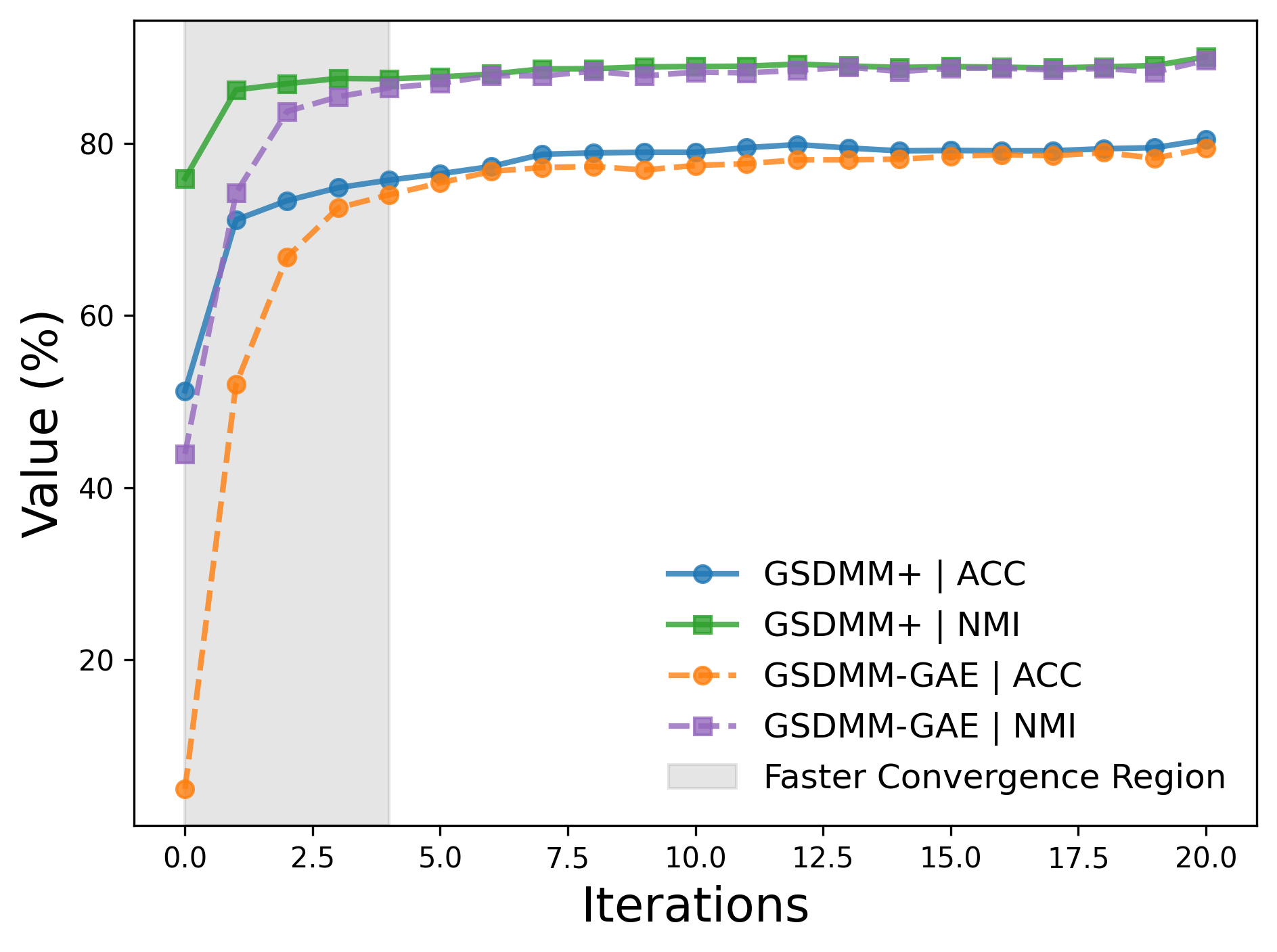}}
\centerline{(a) Tweet}
\end{minipage}
\begin{minipage}{0.49\linewidth}
\centerline{\includegraphics[width=0.98\textwidth]{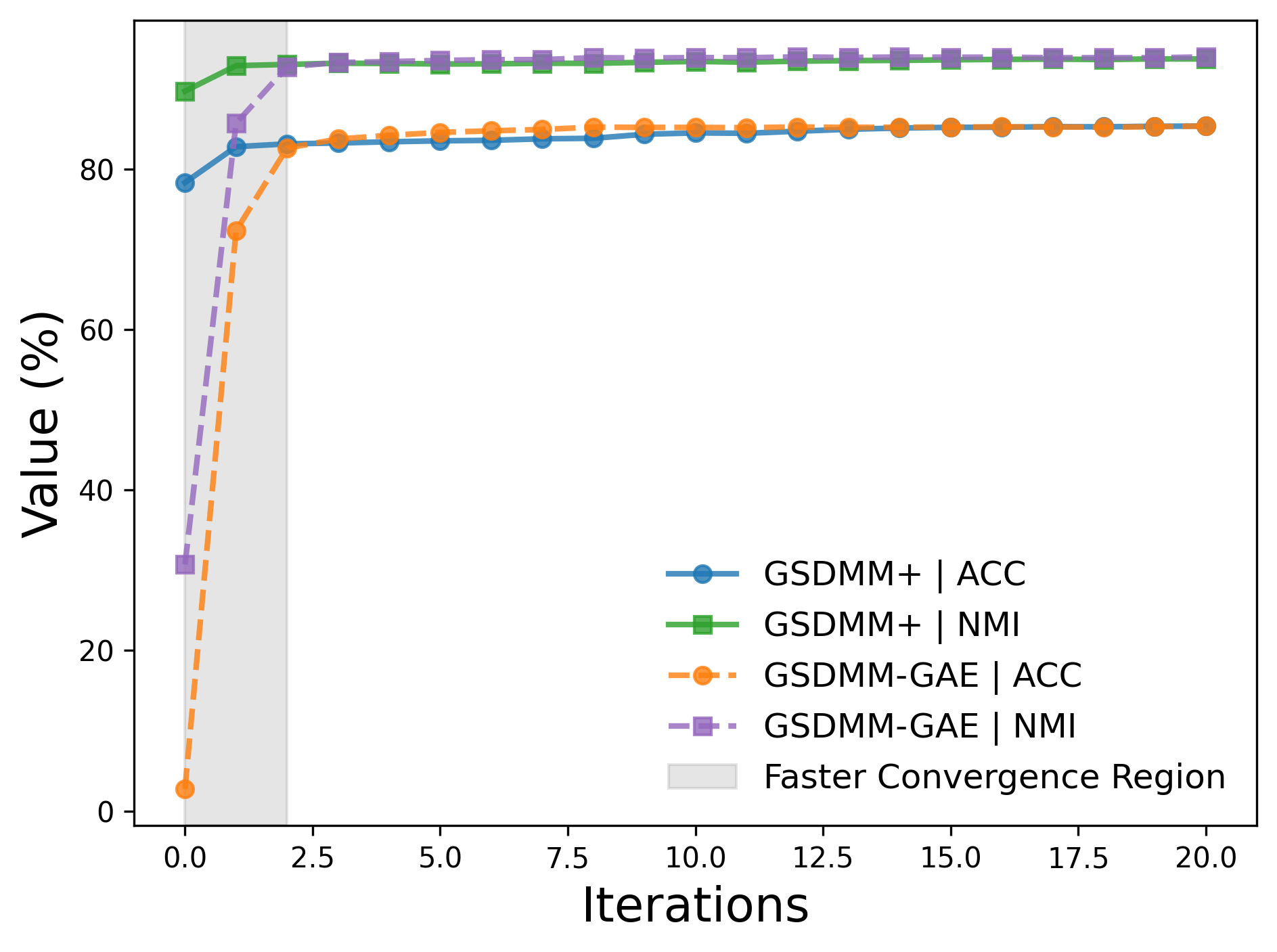}}
\centerline{(b) News-TS}
\end{minipage}
\caption{Comparison of clustering performance using adaptive clustering initialization (\sysname{}) and random initialization (GSDMM-GAE) on the Tweet and News-TS Datasets.}
\label{tweet+TS_initK}
\end{figure}

Figure~\ref{tweet+TS_initK} illustrates the comparative performance of adaptive clustering initialization and random initialization on the Tweet and News-TS datasets, respectively. The results highlight several key observations: 1) Adaptive clustering initialization consistently achieves higher ACC and NMI scores within fewer iterations than random initialization. This indicates that our method facilitates more efficient convergence, reducing the number of iterations required to reach stable and optimal clustering results. 2) Unlike random initialization, which shows significant fluctuations in ACC and NMI scores during the early stages, adaptive clustering initialization maintains a steadier improvement trajectory. It also demonstrates greater stability across iterations, crucial for ensuring reliable clustering outcomes.

The comparison clearly demonstrates that adaptive clustering initialization enhances the efficiency of the clustering process. The \sysname{} algorithm achieves faster convergence and improves performance in early iterations by leveraging a structured initialization strategy. 


\subsubsection{Entropy-based Word Weighting}
Unlike the original DMM model, which assigns equal importance to all words and struggles to adjust the hyper-parameter $\beta$ to obtain an appropriate number of clusters, our method leverages word entropy to represent word importance in clustering. This enables the model to generate finer-grained clusters and enhances its ability to capture deeper insights from textual content.

\begin{table}[htbp]
\caption{Top 20 (left) and bottom 20 (right) items by entropy on Tweet.}
\begin{minipage}{0.45\linewidth}
    \centering
    \renewcommand{\arraystretch}{1.1} 
    \setlength{\tabcolsep}{7pt} 
    \label{tab:top20}
    \begin{tabular}{ccc}
    \toprule
    \textbf{Rank} & \textbf{Word} & \textbf{Entropy} \\
    \midrule
    1  & news     & 1.0000 \\
    2  & will     & 0.8436 \\
    3  & time     & 0.8415 \\
    4  & today    & 0.8128 \\
    5  & day      & 0.7446 \\
    6  & report   & 0.7335 \\
    7  & blog     & 0.7280 \\
    8  & post     & 0.7271 \\
    9  & ap       & 0.7121 \\
    10 & year     & 0.7097 \\
    11 & well     & 0.7084 \\
    12 & good     & 0.7002 \\
    13 & reuters  & 0.6985 \\
    14 & state    & 0.6903 \\
    15 & help     & 0.6730 \\
    16 & going    & 0.6683 \\
    17 & great    & 0.6676 \\
    18 & set      & 0.6670 \\
    19 & read     & 0.6605 \\
    20 & gt       & 0.6576 \\
    \bottomrule
    \end{tabular}
\end{minipage}
\hspace{0.4cm}
\begin{minipage}{0.45\linewidth}
    \centering
    \renewcommand{\arraystretch}{1.1} 
    \setlength{\tabcolsep}{7pt} 
    \label{tab:bottom20}
    \begin{tabular}{ccc}
    \toprule
    \textbf{Rank} & \textbf{Word} & \textbf{Entropy} \\
    \midrule
    1  & debt        & 2.92e-7 \\
    2  & djokovic    & 2.92e-7 \\
    3  & setting     & 2.92e-7 \\
    4  & emanuel     & 2.92e-7 \\
    5  & starbucks   & 2.92e-7 \\
    6  & rahm        & 2.92e-7 \\
    7  & kardashian  & 2.92e-7 \\
    8  & aid         & 2.92e-7 \\
    9  & iran        & 2.92e-7 \\
    10 & chrysler    & 2.92e-7 \\
    11 & aquarium    & 2.92e-7 \\
    12 & cup         & 2.92e-7 \\
    13 & meat        & 2.92e-7 \\
    14 & obamacare   & 2.92e-7 \\
    15 & eminem      & 2.92e-7 \\
    16 & ballot      & 2.92e-7 \\
    17 & chavez      & 2.92e-7 \\
    18 & swan        & 2.92e-7 \\
    19 & olbermann   & 2.92e-7 \\
    20 & term        & 2.92e-7 \\
    \bottomrule
    \end{tabular}
\end{minipage}
\label{tab:two_side_by_side}
\end{table}

As shown in Table~\ref{tab:two_side_by_side}, the following observations can be made: 1) In our method, although most words retain their information after stop words are removed, experimental analysis reveals significant differences in word importance for cluster formation and variations in the amount of information carried by different words.
2) Words with high entropy exhibit significant non-thematic characteristics, indicating that they are broadly distributed across categories and thus less informative for clustering. Their inclusion may introduce noise, highlighting the importance of identifying and down-weighting such words. In contrast, words with entropy close to zero demonstrate strong thematic relevance and serve as reliable indicators of category membership, making them critical for clustering tasks. Therefore, using entropy is essential for evaluating word-topic relevance.

The visualization of entropy confirms that word importance is not uniform. By leveraging word entropy, we can prioritize low-entropy words for clustering while reducing the impact of high-entropy words. This refinement improves the performance of the GSDMM model, leading to more accurate and meaningful clustering results.

\subsubsection{Granularity Adjustment with Cluster Merging}
In our method, entropy-based word weighting enables the discovery of finer-grained clusters and the uncovering of deeper latent topics, which more accurately capture the inherent semantic structure of the data. However, the mismatch between the number of clusters and the true categories negatively impacts performance. 

To address this issue, we apply granularity adjustment with cluster merging. Specifically, clusters with high similarity are merged, reducing redundancy and progressively refining the clustering structure to better align with the true underlying categories. This technique ensures cluster number alignment while preserving fine-grained information extraction, ultimately achieving superior clustering performance.

\begin{figure}[!t]
\centering
\begin{minipage}{0.49\linewidth}
\centerline{\includegraphics[width=0.98\textwidth]{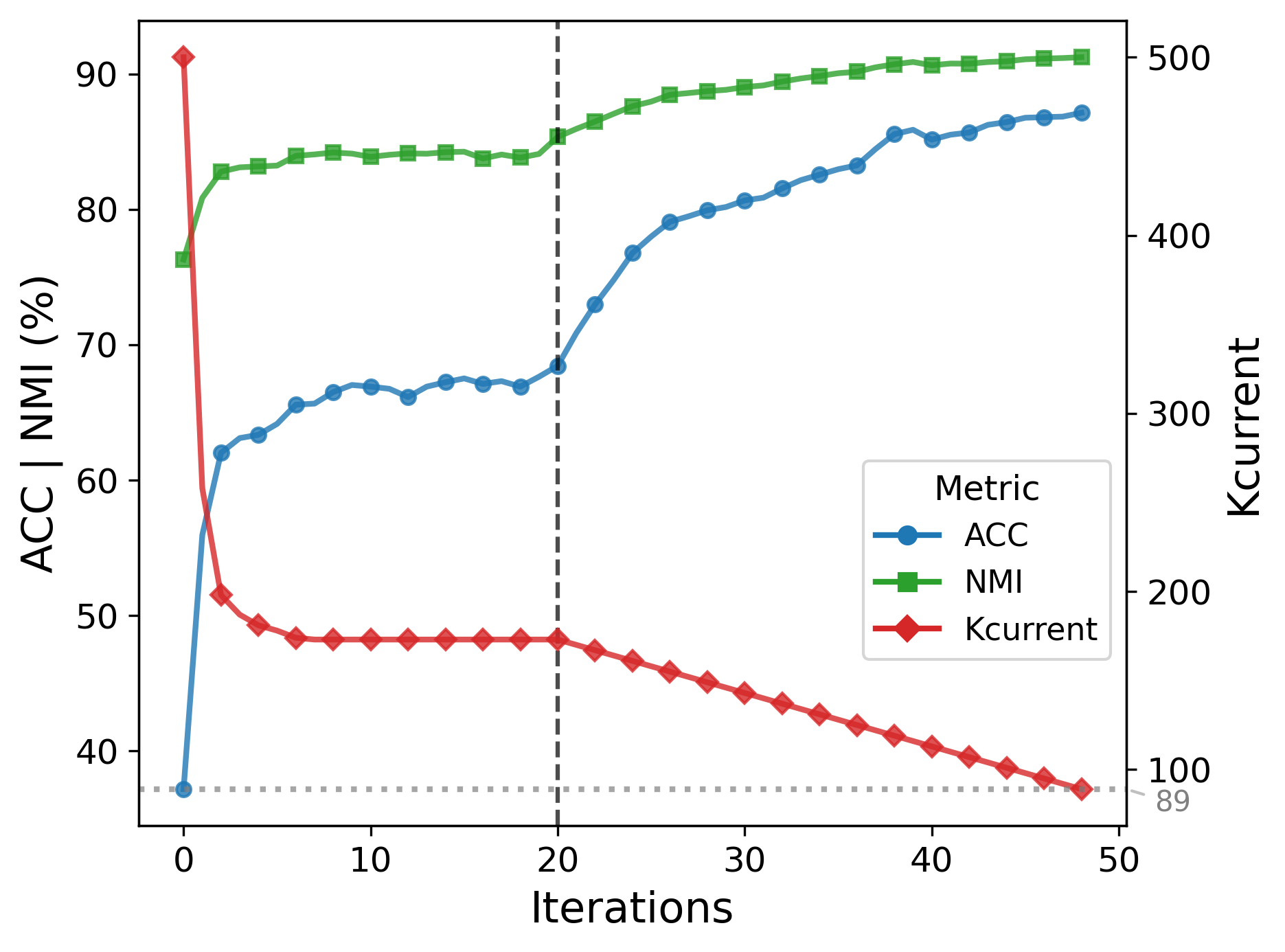}}
\centerline{(a) Tweet}
\centerline{\includegraphics[width=0.98\textwidth]{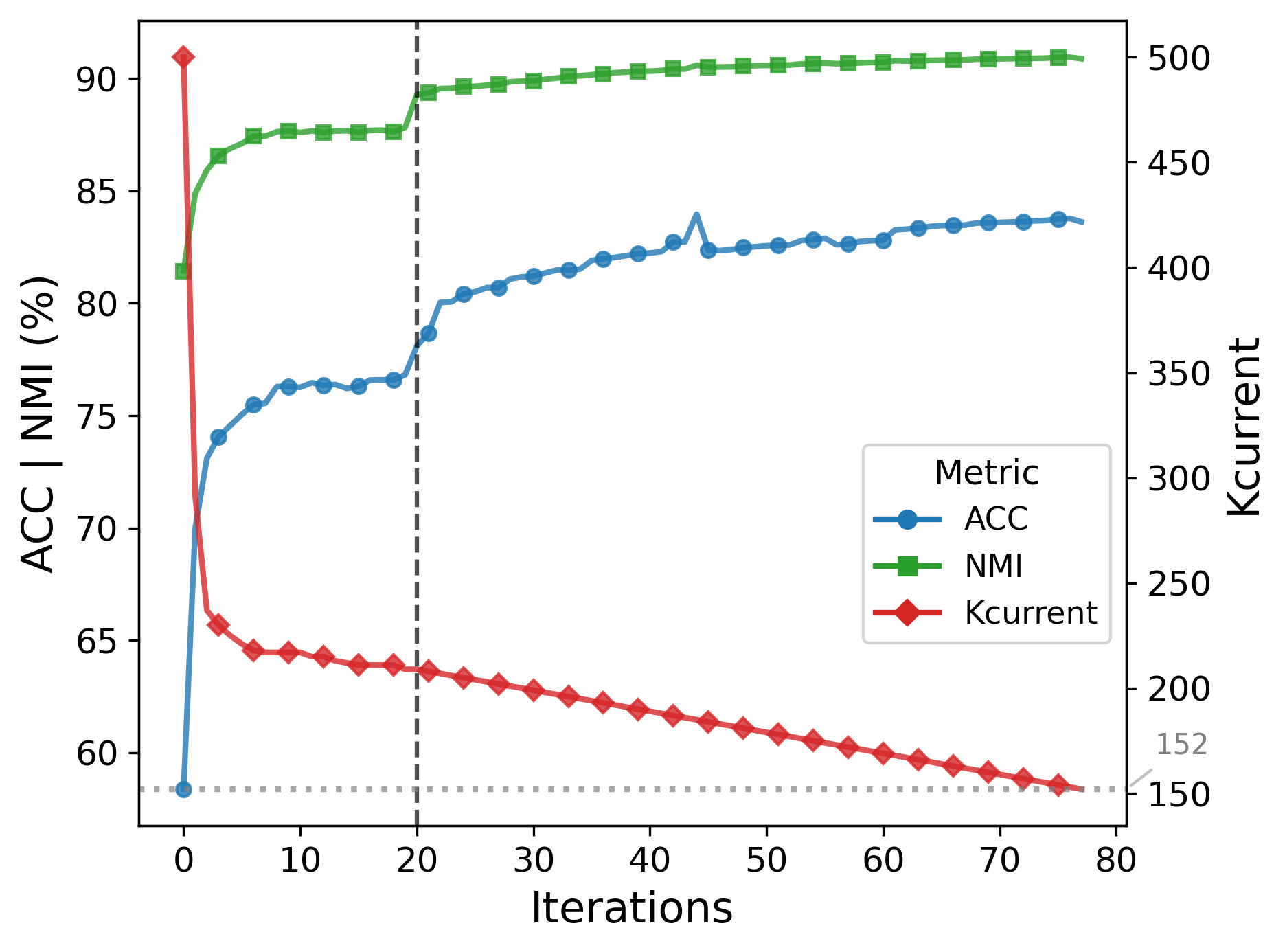}}
\centerline{(c) News-T}
\end{minipage}
\begin{minipage}{0.49\linewidth}
\centerline{\includegraphics[width=0.98\textwidth]{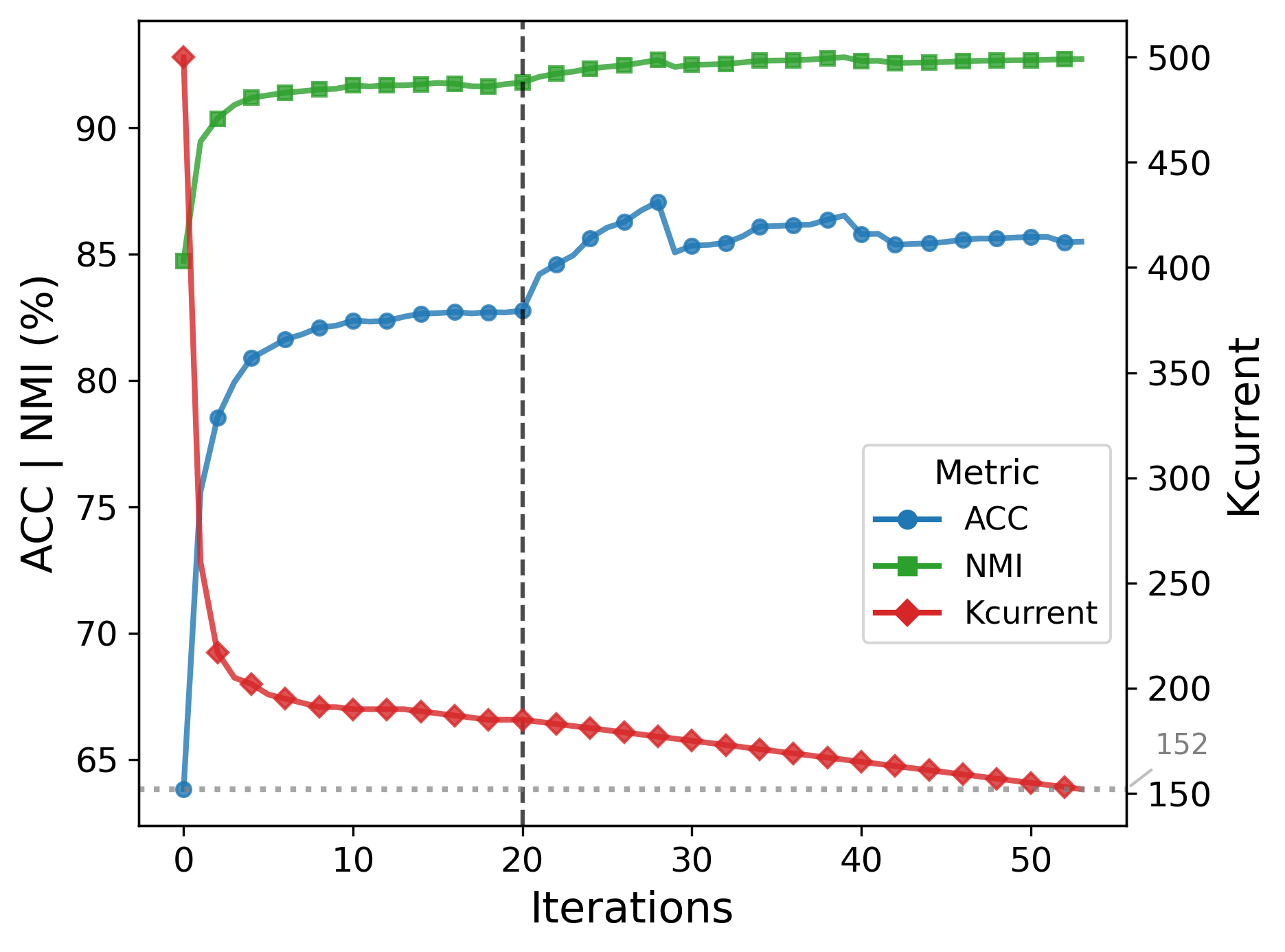}}
\centerline{(b) News-S}
\centerline{\includegraphics[width=0.98\textwidth]{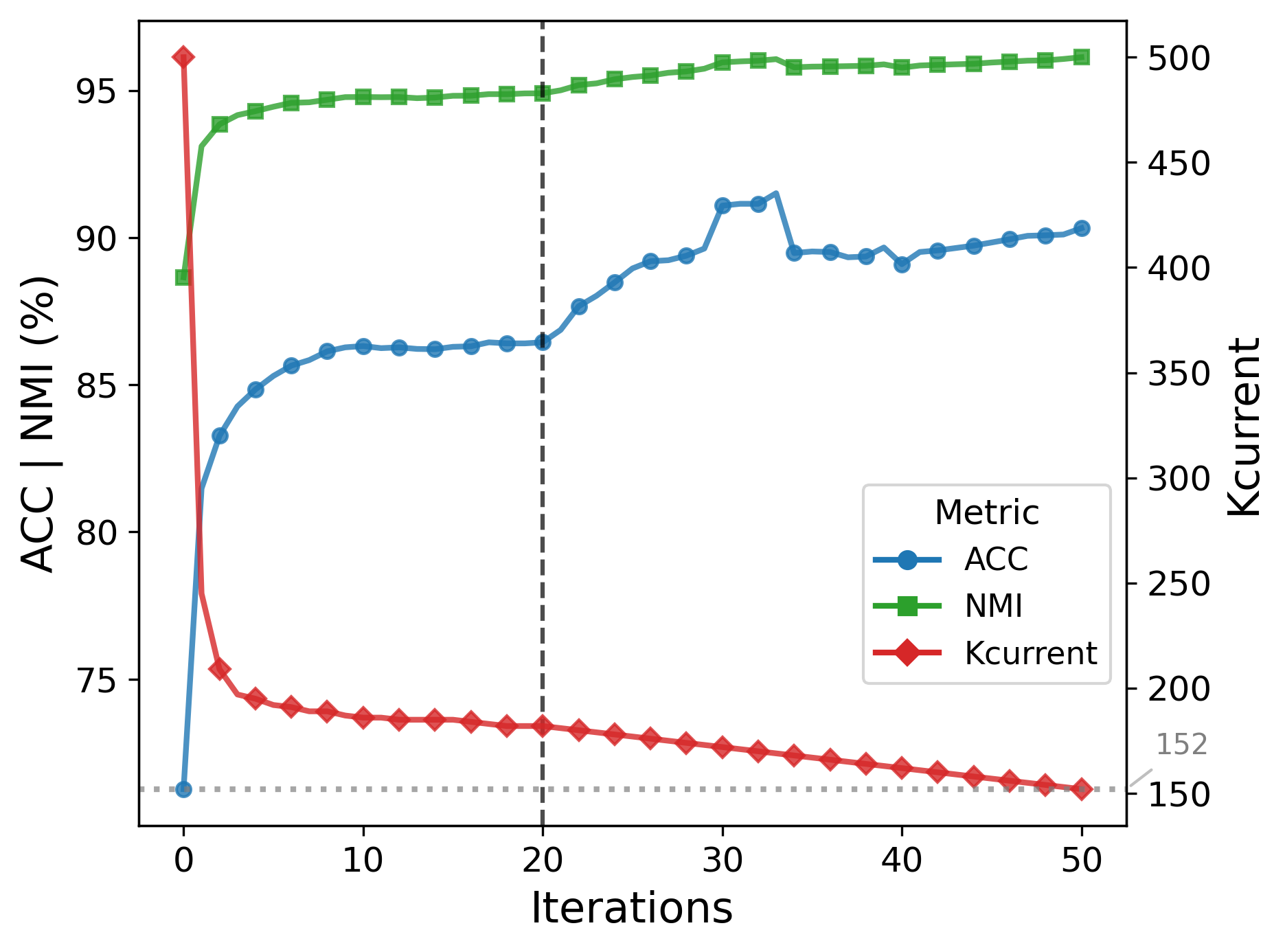}}
\centerline{(d) News-TS}
\end{minipage}
\caption{The performance of \sysname{} over the iterations across four datasets, with vertical dashed lines in the figures indicating the beginning of granularity adjustment.}
\label{png_tfidf}
\end{figure}

Figure~\ref{png_tfidf} illustrates the clustering performance and the variation in the number of clusters across four datasets. The key advantages of this approach are twofold:
1) By iteratively refining the clusters through entropy-based updates, the model obtains more representative word distributions for each cluster. Although this leads to better separation of fine-grained semantic differences between clusters, it causes a deviation between the generated cluster distribution and the true cluster distribution. This design effectively adjusts the cluster granularity and eliminates the gap from the true number of clusters, delivering more accurate and meaningful clustering outcomes. 
2) Experimental results confirm that cluster merging significantly enhances the clustering performance. Compared to origin clustering approaches, cluster merging achieves a better balance between granularity and accuracy, leading to higher ACC and NMI scores.

\subsection{Running Time Analysis}

\begin{table}[htbp]
\caption{Comparison of running time(s) on the Tweet and News-TS datasets.}
    \centering
    \renewcommand{\arraystretch}{1.2} 
    \setlength{\tabcolsep}{9pt} 
    \label{runing_time}
    \begin{tabular}{ccc}
    \toprule
    \textbf{Method} & \textbf{Tweet} & \textbf{News-TS} \\
    \midrule
    GSDMM  & 40     & 606 \\
    GSDMM (w/o Entropy)  & 106     & 1341 \\
    \sysname{}  & 158     & 1392 \\
    MVC  & 447     & 3076 \\
    SCCL  & 1125    & 1622 \\
    DACL  & 2000+    & 3000+ \\
    RSTC  & 3300+    & 4500+ \\
    \bottomrule
    \end{tabular}
\end{table}
In this section, we analyze the running time of our methods on Tweet and News-TS datasets. All experiments are conducted on a single NVIDIA A100 GPU. For all models, we present the average execution time over ten runs. To ensure a fair comparison, the parameters of all models are set according to their original papers. 

Table~\ref{runing_time} presents the experimental results, leading to the following observations: 1) The proposed GSDMM exhibits exceptional computational efficiency across datasets of varying scales. 2) We focus on entropy-based word weighting, which plays a crucial role in extracting the semantic information of words. Comparing the running time of \sysname{} with and without entropy, we find that entropy-based word weighting significantly improves performance with minimal additional computational cost. This further demonstrates its efficiency and effectiveness. 3) Comparing the running time of \sysname{} with the competitive models, we observe that SCCL is less efficient on the Tweet dataset, while MVC is less efficient on the News-TS dataset. DACL and RSTC require too much time to converge, resulting in very low efficiency. In contrast, our model achieves the highest efficiency across both datasets while maintaining strong clustering performance. These results further highlight the superiority of our approach.

\subsection{Hyper-parameter Analysis} \label{Hyper-parameter_Analysis}


\begin{figure}[t]
\centering
\begin{minipage}{0.49\linewidth}
\centerline{\includegraphics[width=0.98\textwidth]{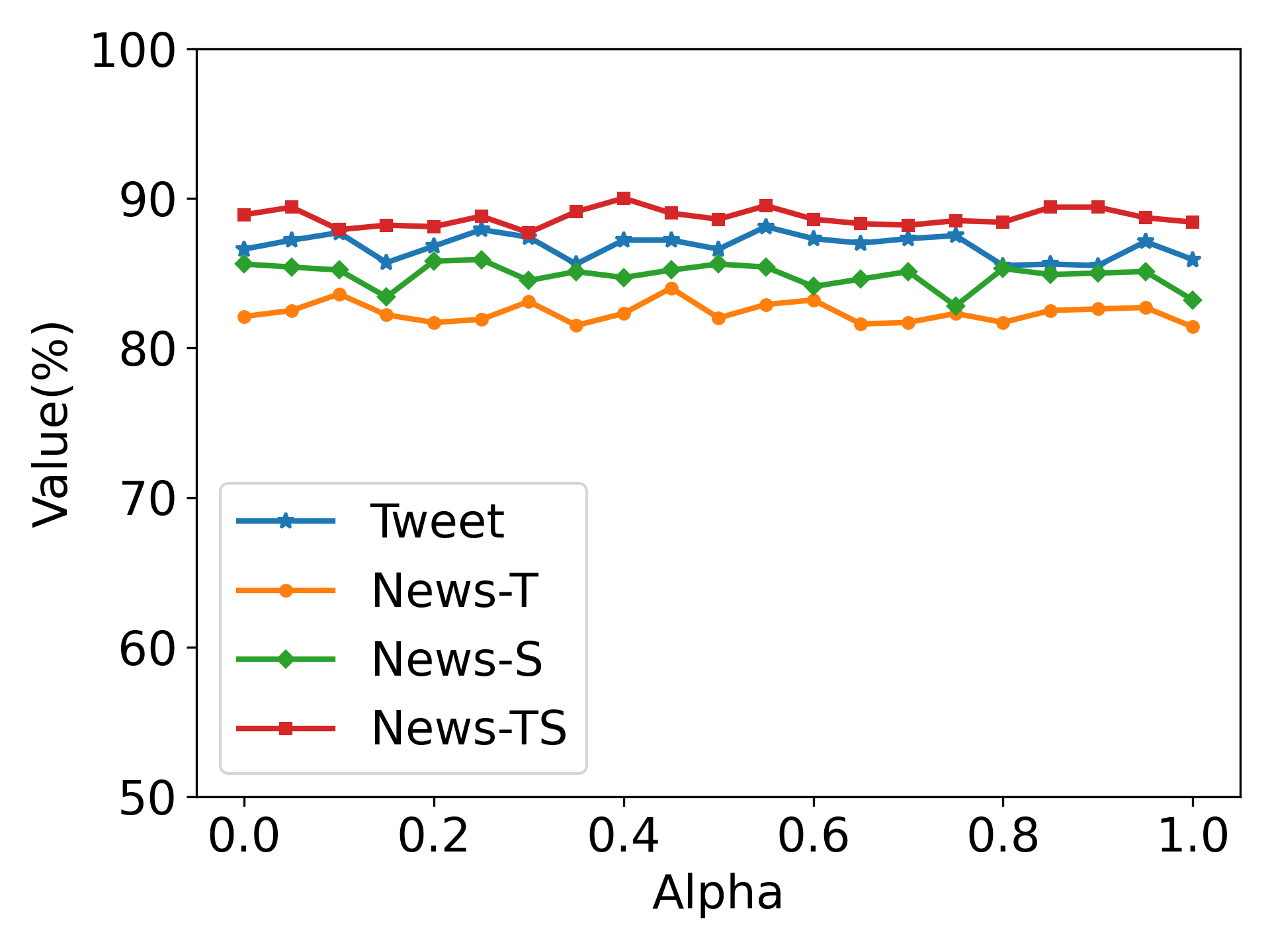}}
\centerline{(a) ACC}
\end{minipage}
\begin{minipage}{0.49\linewidth}
\centerline{\includegraphics[width=0.98\textwidth]{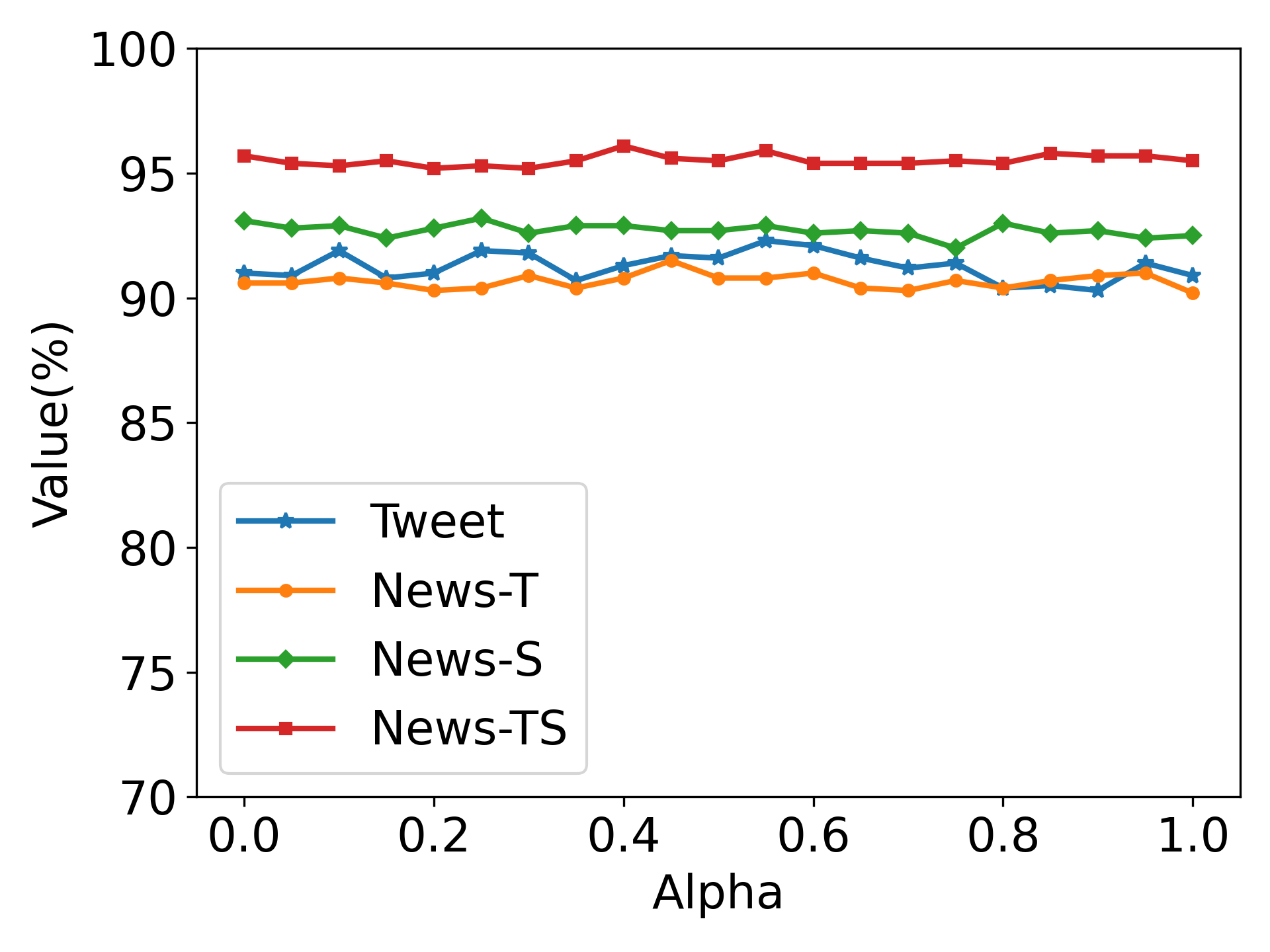}}
\centerline{(b) NMI}
\end{minipage}
\caption{The performance of \sysname{} with different values of $\alpha$ on four datasets.}
\label{alpha_analysis}
\end{figure}
In this section, we primarily present the hyper-parameter analysis of the improved \sysname{}, as it achieves the best performance.

1) \textbf{Influence of $\boldsymbol{\alpha}$:}
We investigate the impact of parameter $\alpha$ and vary the value of $\alpha$ from 0 to 1.0 with a step size of 0.05. The experimental results are presented in Figure~\ref{alpha_analysis}. As illustrated in the figures, we can see the performance of \sysname{} is stable with different values of $\alpha$.

We conduct experiments to investigate the performance of \sysname{} when $\alpha=0$, and find that \sysname{} almost does not result in clusters with only one document when $\alpha=0$. We should note that we can also speed up \sysname{} by setting $\alpha$ to zero. Because when $\alpha=0$, we can just discard clusters that get empty as they will never be chosen again. Futhermore, we should note that Equation~\eqref{eq:dpm3.4.7} cannot be derived from DMM when $\alpha=0$, because $\alpha$ is the parameter of the Dirichlet distribution in DMM and must be larger than zero.

\begin{figure}[!t]
\centering
\begin{minipage}{0.49\linewidth}
\centerline{\includegraphics[width=0.98\textwidth]{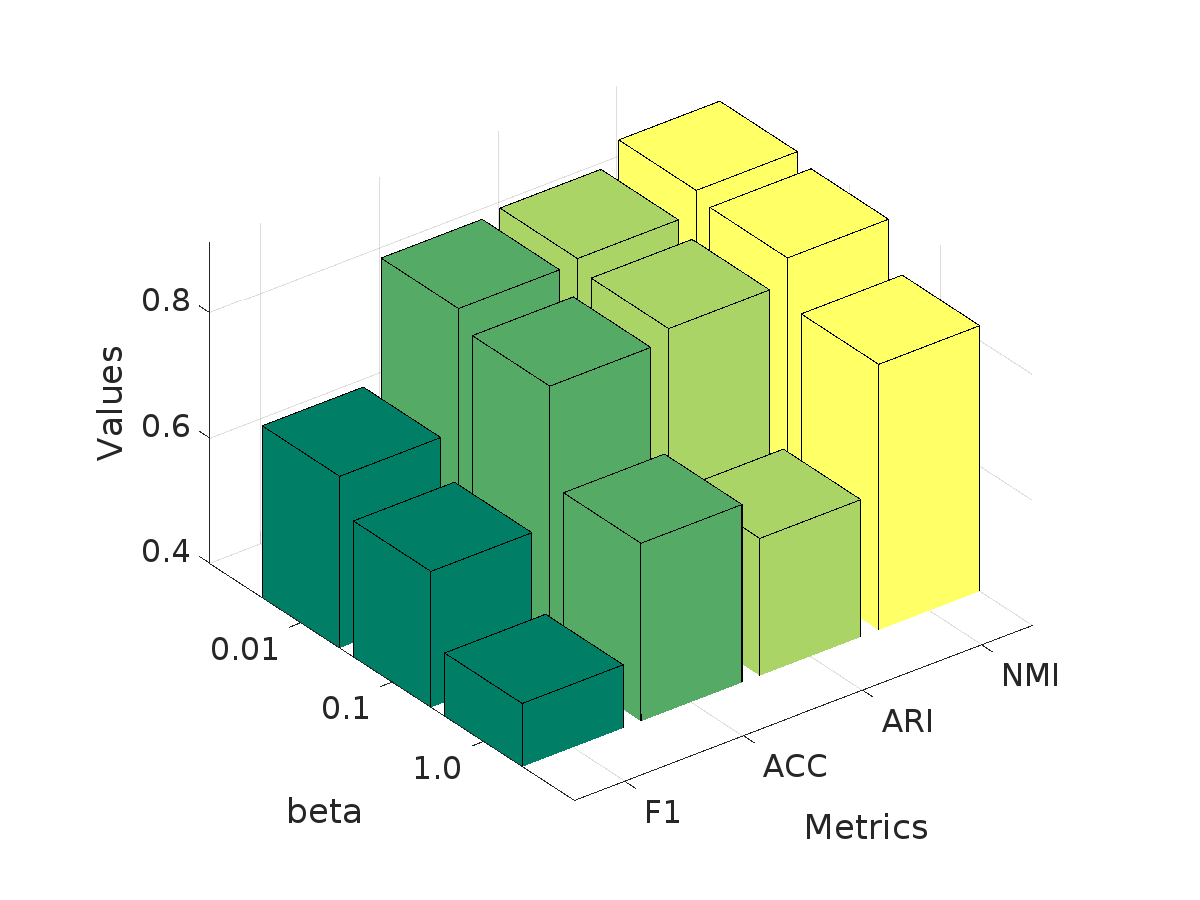}}
\centerline{(a) Tweet}
\centerline{\includegraphics[width=0.98\textwidth]{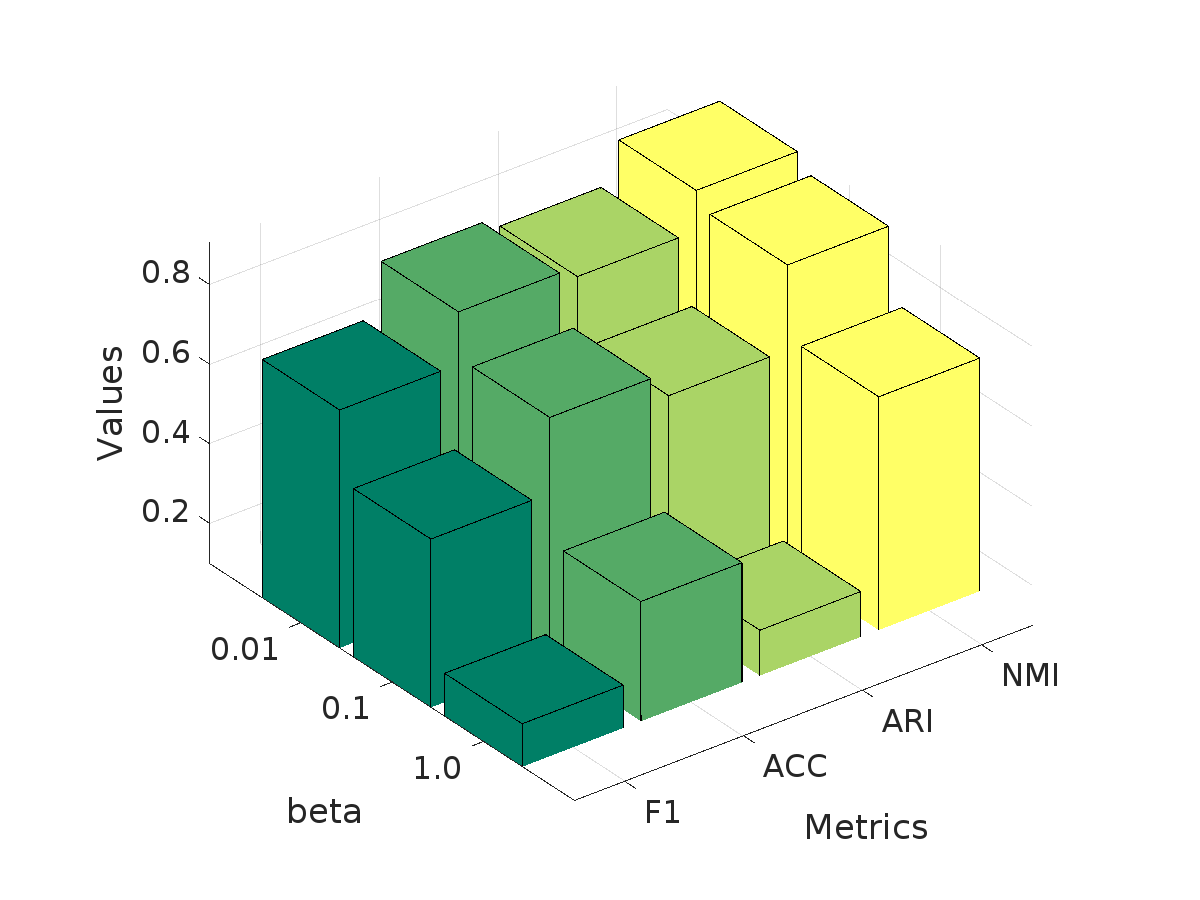}}
\centerline{(c) News-T}
\end{minipage}
\begin{minipage}{0.49\linewidth}
\centerline{\includegraphics[width=0.98\textwidth]{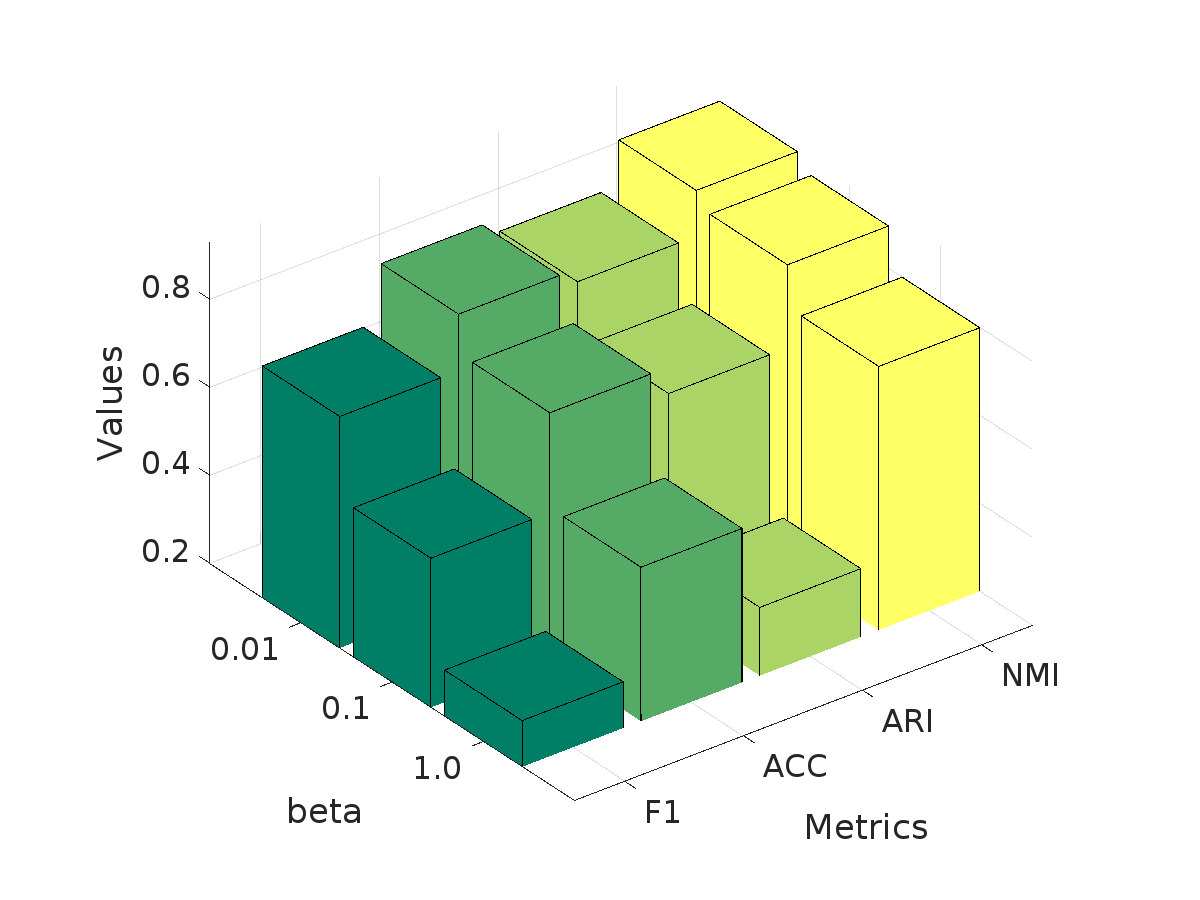}}
\centerline{(b) News-S}
\centerline{\includegraphics[width=0.98\textwidth]{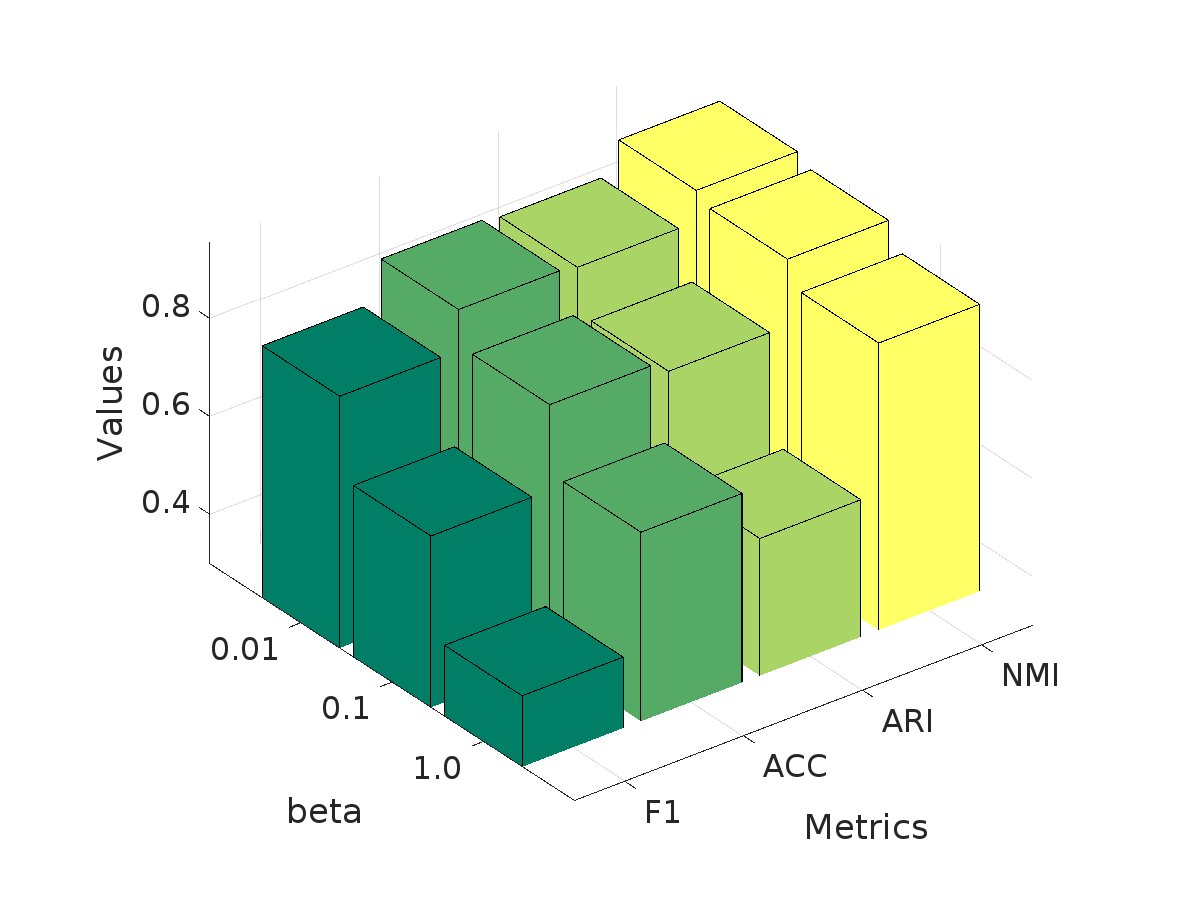}}
\centerline{(d) News-TS}
\end{minipage}
\caption{Hyper-parameter analysis for $\beta$.}
\label{hyper_parameters_beta}
\end{figure}

2) \textbf{Influence of $\boldsymbol{\beta}$:}
We try to investigate the influence of $\beta$ to the number of clusters found by \sysname{} and the performance of \sysname{}.


\begin{table}[ht]
\centering
\caption{The accuracy and NMI of \sysname{} with different $\beta$ values used in adaptive clustering initialization.}
\renewcommand{\arraystretch}{1.3} 
\setlength{\tabcolsep}{6pt} 
\begin{tabular}{c|c|cccccc}
\toprule
\textbf{Dataset} & \textbf{beta ($\beta$)} & 0 & 1e-5 & 1e-4 & 0.001 & 0.01 & 0.1 \\ 
\midrule
\multirow{2}{*}{News-TS} 
& Accuracy (\%) & fail & 89.6 & 89.5 & 89.5 & 88.9 & 82.4 \\
& NMI (\%) & fail & 95.8 & 95.9 & 96.0 & 95.7 & 93.1 \\
\midrule
\multirow{2}{*}{News-S}  
& Accuracy (\%) & fail & 82.5 & 81.3 & 83.8 & 85.5 & 76.4 \\
& NMI (\%) & fail & 92.3 & 91.9 & 92.6 & 92.9 & 89.2 \\
\bottomrule
\end{tabular}
\label{beta_analysis}
\end{table}

\begin{figure}[t]
\centering
\begin{minipage}{0.98\linewidth}
\centerline{\includegraphics[width=0.98\textwidth]{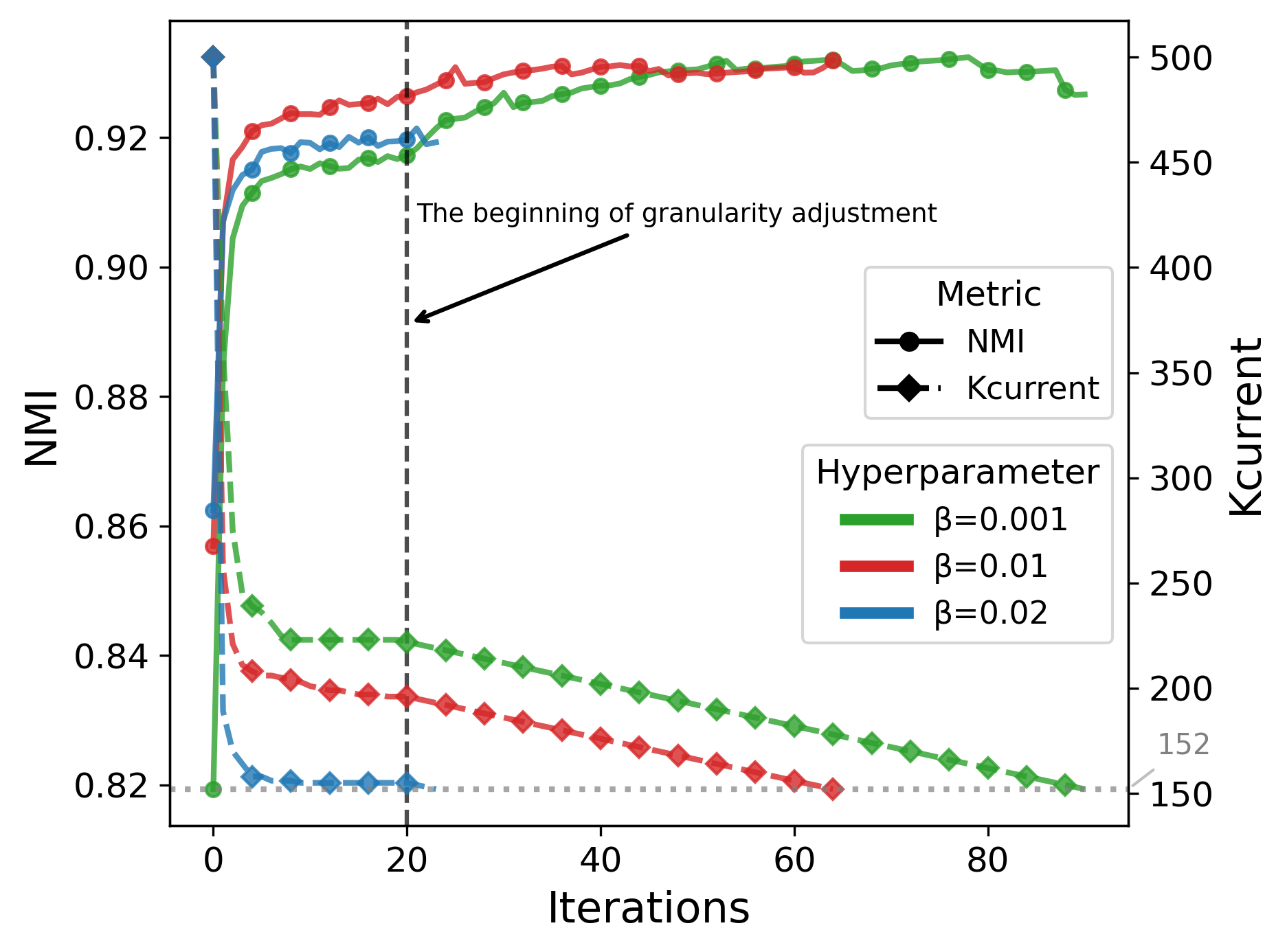}}
\end{minipage}
\caption{The variations in NMI and the number of clusters on the News-S dataset.}
\label{S_deep_beta}
\end{figure}

As shown in Table~\ref{beta_analysis}, in the News-TS dataset, our model performs reasonably well when $\beta$ is in the range of $1 \times 10^{-5}$ to 0.01. However, in the News-S dataset, as $\beta$ decreases, the granularity becomes excessively fine, leading to ineffective cluster merging and a decline in clustering performance. Additionally, when $\beta$ is too large (e.g., 0.1), the accuracy and NMI decrease significantly. At the extreme case of $\beta = 0$, as discussed in Section~\ref{sec:disAlpahAndBeta}, the model cannot be trained. Figure~\ref{S_deep_beta} provides a clearer illustration of how NMI and the number of clusters change with $\beta$ on the News-S dataset. These results validate our approach. By choosing $\beta = 0.01$, we achieve sufficiently fine-grained clusters that capture subtle local patterns while also benefiting from effective cluster merging for improved clustering performance.

\subsection{Visualization Analysis} \label{Visualization_Analysis}

\begin{figure*}[t]  
    \centering
    \begin{minipage}[t]{0.23\textwidth}
        \centering
        \includegraphics[width=\linewidth]{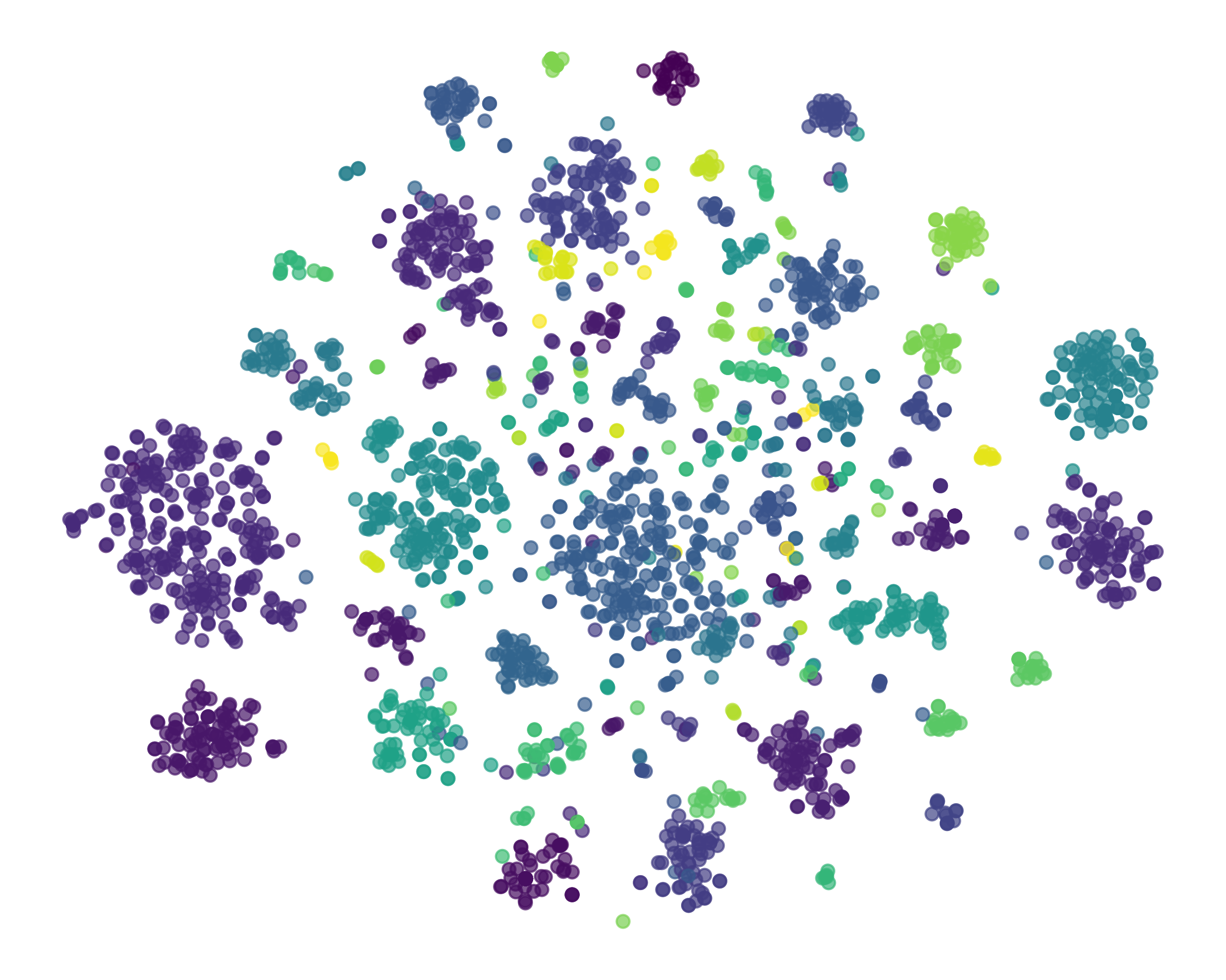}
        \centerline{(a) Tweet}
        \label{fig:1}
    \end{minipage}
    \hfill
    \begin{minipage}[t]{0.23\textwidth}
        \centering
        \includegraphics[width=\linewidth]{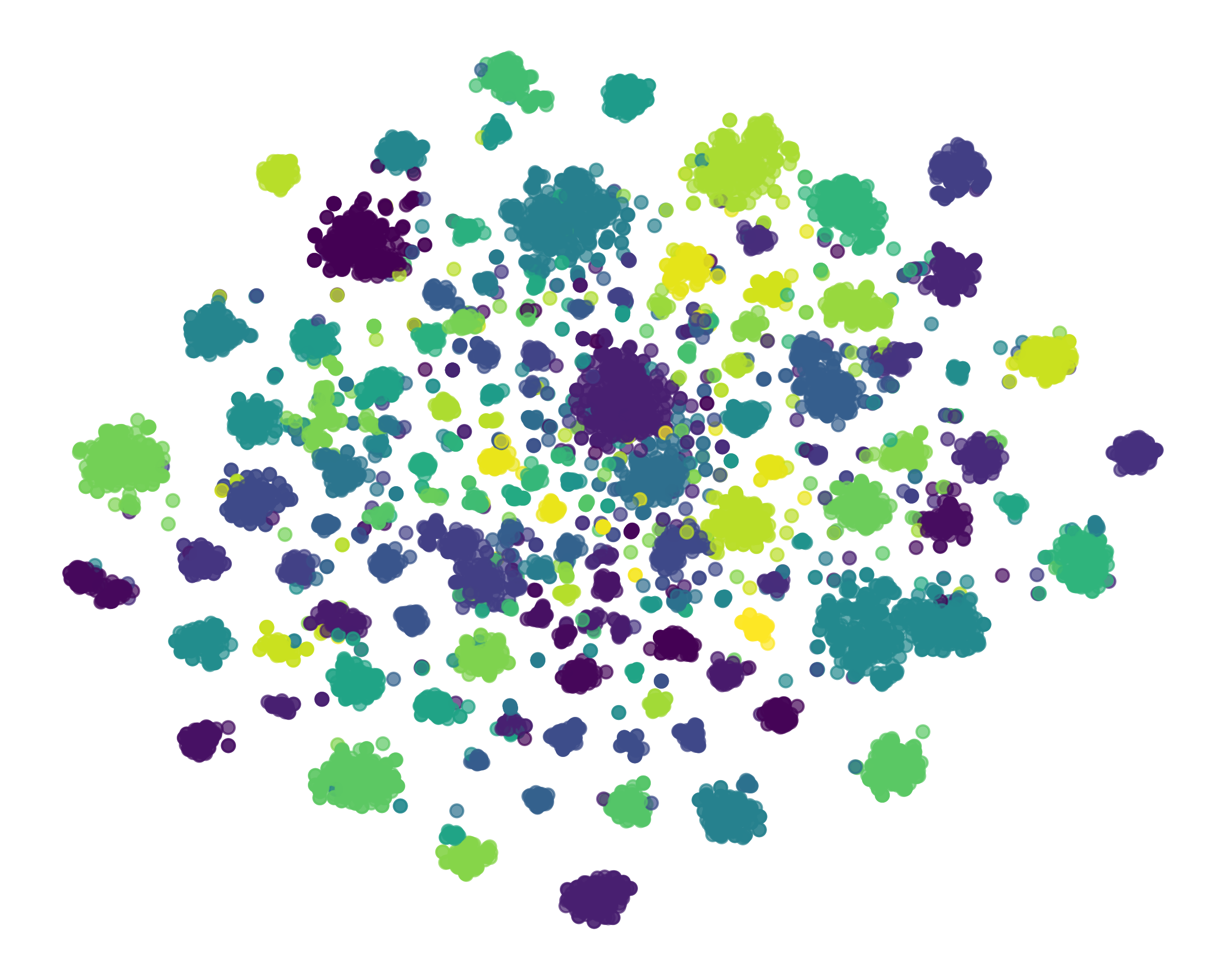}
        \centerline{(b) News-S}
        \label{fig:2}
    \end{minipage}
    \hfill
    \begin{minipage}[t]{0.23\textwidth}
        \centering
        \includegraphics[width=\linewidth]{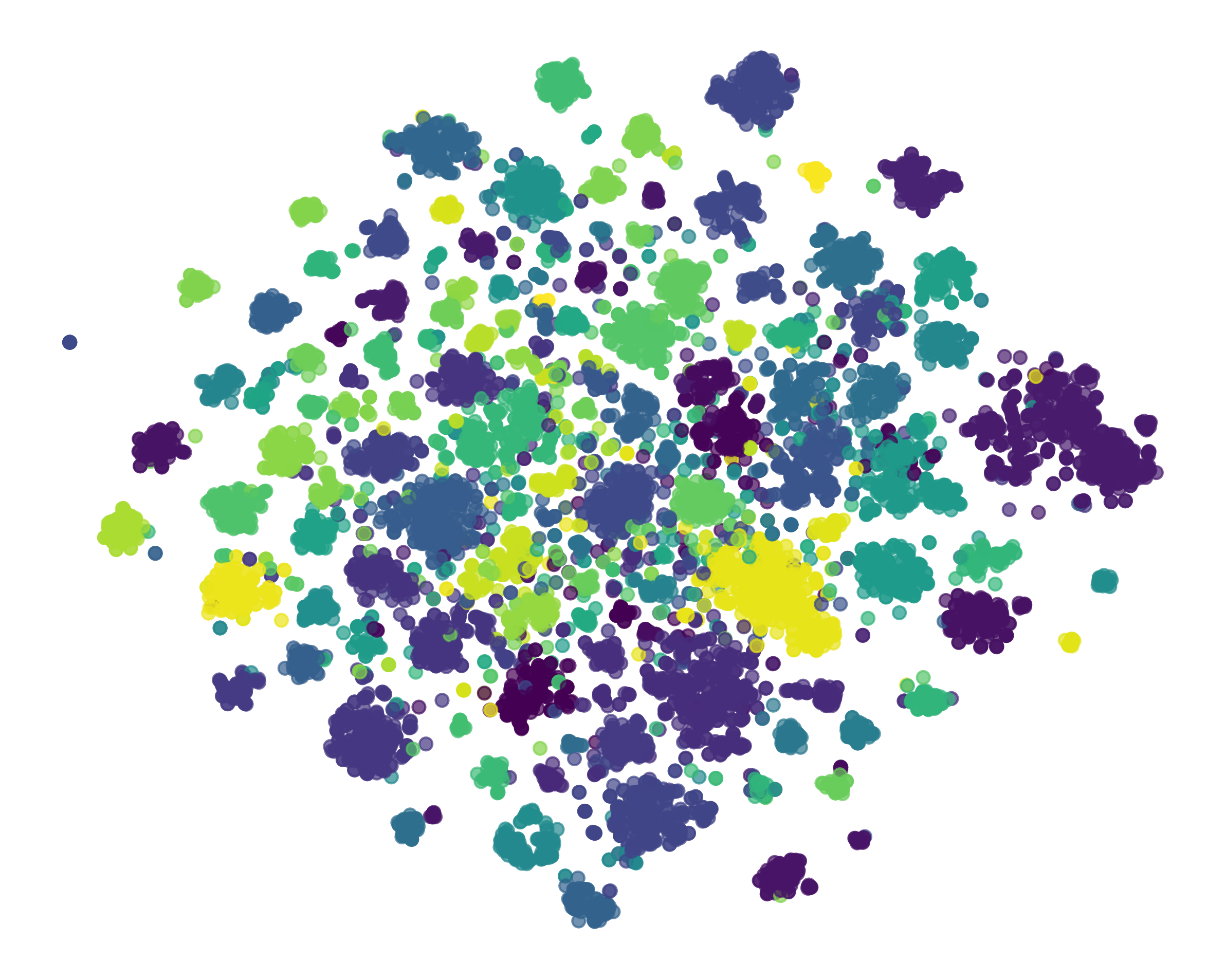}
        \centerline{(c) News-T}
        \label{fig:3}
    \end{minipage}
    \hfill
    \begin{minipage}[t]{0.23\textwidth}
        \centering
        \includegraphics[width=\linewidth]{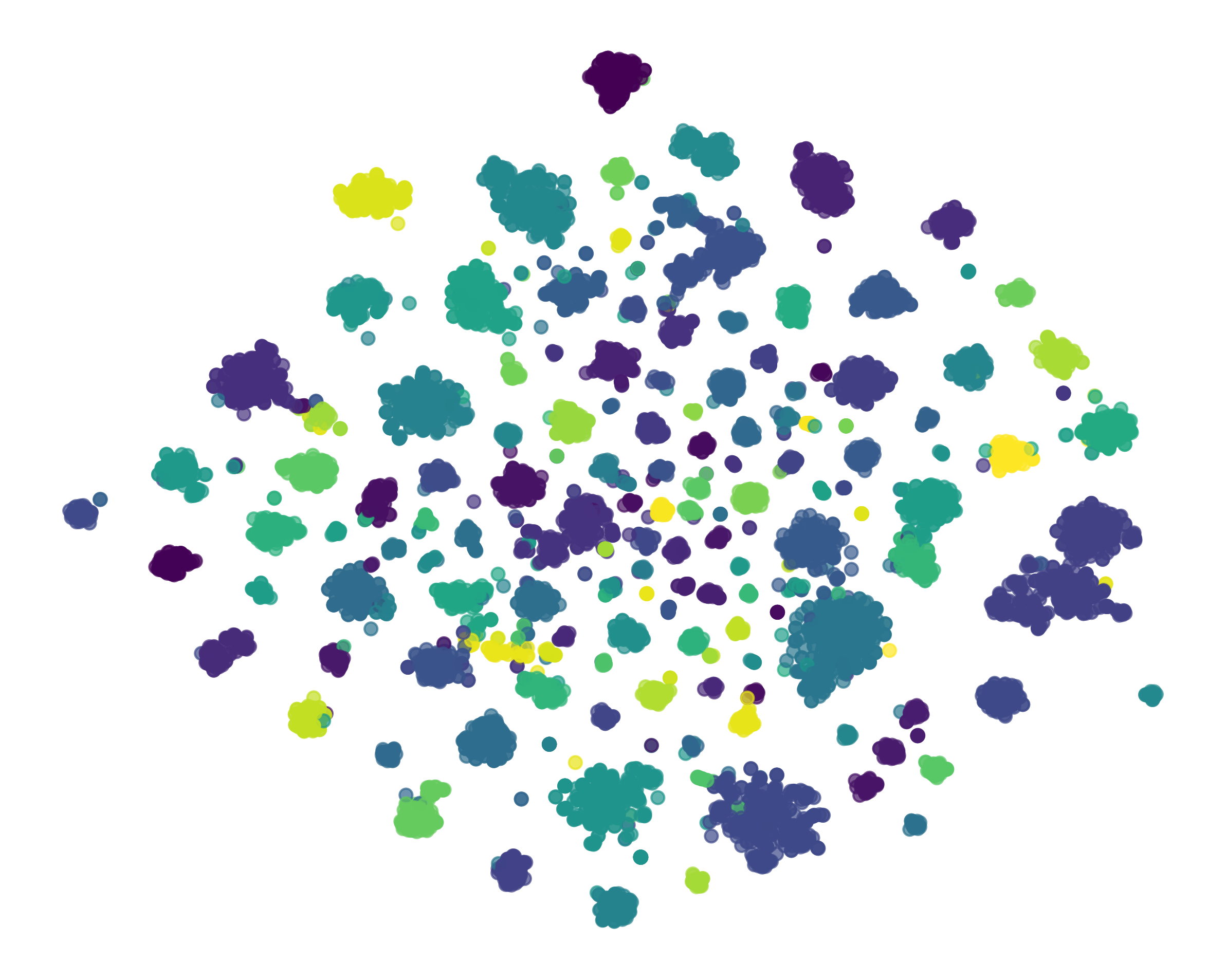}
        \centerline{(d) News-TS}
        \label{fig:4}
    \end{minipage}
\caption{2D t-SNE visualization of \sysname{} on four benchmark datasets}
\label{tsne_all}
\end{figure*}

\begin{table*}[htbp]
\centering
\renewcommand{\arraystretch}{1.2} 
\setlength{\tabcolsep}{7pt} 
\caption{The top ten representative words for the first 18 clusters discovered by \sysname{} on the News-TS dataset.}
\label{tab:clusters}
\begin{tabular}{|l|l|l|l|l|l|l|l|l|l|}
\hline
\textbf{Cluster 1} & \textbf{Cluster 2} & \textbf{Cluster 3} & \textbf{Cluster 4} & \textbf{Cluster 5} & \textbf{Cluster 6} & \textbf{Cluster} & \textbf{Cluster 8} & \textbf{Cluster 9} \\
\hline
berlusconi  & typhoon	  & google	   & illinois	 & texas	 	   & kanye	    & blackberry	 & xbox	    & hpv	 \\
vote	   & philippine	 & search	  & northern	 & nurse	 	   & west	    & bbm	         & microsoft&	vaccine	 \\
silvio	   & haiyan	     & chrome	  & lynch	     & stabbing	 	   & kim	    & android	     & console	& cancer	\\
senate	   & relief	   & voice	      & michigan	 & hospital	 	   & kardashian	 & channel	   & game	     & vaccination	\\
italian	   & aid	    & extension	    & western	 & longview	 	   & bound	    & user	      & launch	 & human	\\
italy	   & help	    & browser	    & jordan	 & center	 	   & rogen	    & device	 & playstation	& cervical	\\
rome	    & effort	 & hotword	    & husky	     & good	     	   & video	    & social	 & disc	 &papillomavirus	\\
minister	& manila	& desktop	   & game	     & tuesday	 	   & franco	    & messaging	 & user	 &gardasil	\\
government	& tacloban & hand	     & niu	         & shepherd	 	   & seth	    & app	     & drive	&study	 \\
prime	     & nov	  & free         & quarterback	 & medical	 	   & james	    & messenger	 & sony	&hea \\
\hline
\textbf{Cluster 10} & \textbf{Cluster 11} & \textbf{Cluster 12} & \textbf{Cluster 13} & \textbf{Cluster 14} & \textbf{Cluster 15} & \textbf{Cluster 16} & \textbf{Cluster 17} & \textbf{Cluster 18}\\
\hline
comet	     & lakers	    & hanukkah	     & hewlett	 & group	      	 & black	    & launch		 & chelsea	    &homefront	\\
ison	     & bryant	    & thanksgiving	 & packard	 & tax	          	 & friday	    & spacex		 & basel		& statham	\\
sun	         & kobe	        & thanksgivukkah & hp	     & political	 	 & thanksgiving	 & satellite	 & league		& jason		\\
nasa	     & los	        & year	         & earnings	 & irs	          	 & shopping	     & rocket		 & champion		&movie		\\
thanksgiving	& angeles	 & holiday	     & quarter	 & exempt	         & deal	        & falcon		 & mourinho		&action	\\
nov	       & extension	    & day	        & hpq	     & rule	             & holiday	    & commercial	 & jose		    &stallone	\\
day	       & wizard	       & jewish	        & company	 & obama	      	 & day	       & monday			 & tuesday		& review	\\
solar	   & year	       & time	        & revenue	 & proposed	      	 & store	   & attempt		 & group		& franco	\\
astronomer	 & contract	   & menorah	    & share	     & tuesday	      	 & monday	   & space		     & salah		& sylvester	\\
image	   & washin	       & turkey	        & fourth	 & administration 	 & cyber	   & cape		     & despite		& film	\\
\hline
\end{tabular}
\end{table*}

To further demonstrate the superiority of \sysname{} intuitively, we conduct 2D t-SNE~\cite{van2008visualizing} on four different datasets. Unlike previous methods that rely on model-learned embeddings and ground-truth labels to evaluate representation learning capabilities, we adopt a different approach in this section. Specifically, we use TF-IDF to obtain fixed vector representations for each text and perform t-SNE visualization in conjunction with the cluster labels obtained from our model. This approach effectively evaluates the quality of the clustering results. As shown in Figure~\ref{tsne_all}, our model successfully groups text data into distinct clusters, demonstrating its effectiveness.

Table~\ref{tab:clusters} presents the top ten representative words for the first 18 clusters discovered by \sysname{} on the News-TS dataset. Since the dataset is categorized based on events, a single representative word may not fully convey the meaning of class. However, by analyzing the representative words within each cluster, we observe that most clusters exhibit strong semantic consistency and clear event-oriented categorization, effectively capturing key information across different domains. For example, cluster 3 pertains to discussions on google search and chrome browser features, while cluster 8 focuses on gaming and console-related topics.


\section{Related Works} \label{related_works}

\subsection{Short Text Clustering Algorithms based on Topic Models}
Topic models assume data points are generated by a mixture model, and then employ techniques like EM~\cite{bishop2006pattern} or Gibbs sampling~\cite{robert2004monte} to estimate the parameters of the mixture model, so as to obtain the clustering results.

Nigam et al.~\cite{nigam2000text} utilize the DMM model for classification with both labeled and unlabeled documents and propose an EM-based algorithm~\cite{nigam2000text} for DMM (abbr. to EM-DMM). When only unlabeled documents are provided, DMM turns out to be a clustering model. Yu et al.~\cite{YuDPM:KDD2010} propose the DPMFS model for text clustering that can also infer the number of clusters automatically. Due to the slow convergence of DPMFS~\cite{DBLP:journals/tkde/HuangYWZS13}, they further propose the DMAFP model as an approximation to the DPMFS and introduce a variational inference algorithm. They compare DMAFP with other four clustering models: EM-DMM~\cite{nigam2000text}, K-means~\cite{jain2010K-means}, LDA~\cite{Blei:LDA2003}, and EDCM~\cite{DCMelkan2006clustering}. Rangrej et al.~\cite{Rangrej:WWW2011} compare three clustering algorithms for short text clustering: K-means, singular value decomposition and affinity propagation on a small set of tweets, and find that Affinity Propagation~\cite{frey2007clustering} can achieve better performance than the other two algorithms. However, the complexity of Affinity Propagation is quadratic in the number of documents, which means Affinity Propagation cannot scale to huge datasets with millions of documents.
Banerjee et al.~\cite{Banerjee:SIGIR2007} propose a method of improving the accuracy of short text clustering by enriching their representation with additional features from Wikipedia. Topic models like LDA~\cite{Blei:LDA2003} and PLSA~\cite{hofmann1999probabilistic} are probabilistic generative models that can model texts and identify latent semantics underlying the text collection. 
LDA has excellent performance on normal-length texts, but it is inferior on short texts because LDA depends on cross-text word co-occurrences, which are lacking in short texts. 
Therefore, we propose GSDMM to address the short text clustering problem and manage the high dimensionality of short texts, achieving excellent performance in clustering short texts. Like Topic Models~\cite{Blei:LDA2003}, GSDMM can also obtain the representative words of each cluster. This enables GSDMM to efficiently address key clustering challenges, including the sparsity issue in short texts. Furthermore, previous methods overlooked the importance of individual words to text clustering, thereby oversimplifying the nature of textual data. In reality, words exhibit varying levels of discriminative power across categories. Therefore, \sysname{} is proposed to adaptively adjust the influence of individual words based on their informational significance within each cluster. This approach highlights discriminative words while minimizing the impact of less informative terms, effectively capturing subtle local patterns.

\subsection{Deep Short Text Clustering Algorithms}
Text clustering has been an enduring research focus. Many early deep text clustering algorithms employ various autoencoder to learn deep representations for clustering. Methods in the former category perform clustering in a two-stage manner, which first feeds texts into a deep neural network to learn the text representations, and then a certain clustering method, like K-means, is performed based on the learned representations. Xu et al~\cite{stcc} propose a self-taught learning framework constituted of a convolutional neural network. Guan et al.~\cite{guan2020deep} use the neural language model and InferSent~\cite{conneau2017supervised} to obtain short text embeddings and then apply K-means for clustering. The two-stage deep clustering methods separate the optimization process into the representations learning step and clustering step, where the objective of representation learning is not consistent with clustering sometimes, leading to unsatisfactory results.

Methods in the latter category perform representation learning and text clustering in an end-to-end manner, thereby allowing joint optimization of the two steps. ARL~\cite{arl} integrates these two components into a unified model that constructs low-dimensional text representations from one-hot encoding through word embedding and mean pooling. Subsequently, attention techniques~\cite{attention} are leveraged between text representations and cluster representations. DECCA~\cite{DECCA} learns text representations using a contrastive autoencoder, upon which a deep embedding clustering framework is built. Guan et al.~\cite{guan2020deep} propose a deep feature-based text clustering model named DFTC, which incorporates text encoders and an explanation module into text clustering tasks. SCCL~\cite{SCCL} develops a novel contrastive learning framework that pulls together the learned representations of positive pairs augmented from the same text while pushing apart negative pairs from other texts. SCCL utilizes the Sentence-BERT~\cite{conneau2017supervised} model and fine-tunes it using contrastive loss and clustering loss to obtain high-quality text embeddings. As is well known, efficiency is extremely important in clustering tasks. 
However, past methods, whether introducing embedding representations or various deep models, have incurred significant efficiency losses while improving effectiveness. Building upon GSDMM and \sysname{}, we have ensured high efficiency while greatly enhancing the performance of short text clustering.

\section{Conclusion} \label{sec:conclusion}
In this paper, we propose a collapsed Gibbs Sampling algorithm for the Dirichlet Multinomial Mixture model (GSDMM) for short text clustering, which effectively addresses sparsity and high dimensionality. GSDMM can also extract representative words for each cluster, enhancing the interpretability of the results. Next, based on several aspects of GSDMM that warrant further refinement, we propose an improved approach, \sysname{}, designed to further optimize its performance. \sysname{} reduces initial noise compared to random initialization by assigning each document to a cluster based on its probability of belonging to each cluster. Furthermore, \sysname{} employs entropy-based word weighting to adaptively adjust word importance and refines clustering granularity through cluster merging to better align the predicted distribution with the true category distribution. This effectively uncovers more topic-related information and enhances clustering performance.
Finally, we conduct extensive experiments, comparing our methods with both classical and state-of-the-art approaches. The experimental results demonstrate the efficiency and effectiveness of our method.
In the future, we will explore how to integrate text embeddings with bag-of-words information to gain a deeper understanding of texts and enhance clustering performance while ensuring high efficiency.


\bibliographystyle{IEEEtran}
\bibliography{Reference}


 




\vfill

\end{document}